\documentclass[11pt]{article}

\usepackage[preprint]{acl}

\usepackage{times}
\usepackage{latexsym}

\usepackage[T1]{fontenc}

\usepackage[utf8]{inputenc}

\usepackage{microtype}

\usepackage{inconsolata}

\usepackage{graphicx}

\usepackage{pifont}
\usepackage{spverbatim}

\usepackage{multirow}
\usepackage[normalem]{ulem}
\useunder{\uline}{\ul}{}
\usepackage{subfigure}
\usepackage[ruled,vlined,linesnumbered]{algorithm2e}
\SetKwComment{Comment}{$\triangleright$\ }{}
\usepackage{graphicx}
\usepackage{caption}
\usepackage{float}
\usepackage[flushleft]{threeparttable}
\usepackage{array,booktabs,makecell}
\usepackage{graphicx}
\usepackage{enumitem}

\usepackage{balance}
\usepackage{footmisc}
\usepackage{tabularx}
\usepackage{tikz}
\usepackage{xcolor}
\usepackage{multirow}
\usepackage[many]{tcolorbox}
\usepackage{subcaption}
\usepackage[export]{adjustbox}
\usepackage{adjustbox}
\usepackage{listings}
\usepackage{xspace}

\usepackage{hyperref}
\newtcolorbox{mybox}[2][]{text width=0.95\linewidth,fontupper=\normalsize,
fonttitle=\bfseries\sffamily\scriptsize, colbacktitle=darkgrey,enhanced,
attach boxed title to top left={yshift=-2mm,xshift=3mm},
boxed title style={sharp corners},top=4pt,bottom=2pt,left=2pt,right=2pt,
  title=#2,colback=white}

\newcommand{\Name}{\texttt{Aletheia}\xspace}

\newenvironment{packeditemize}{
\begin{list}{$\bullet$}{
\setlength{\labelwidth}{6pt}
\setlength{\itemsep}{0pt}
\setlength{\leftmargin}{\labelwidth}
\addtolength{\leftmargin}{\labelsep}
\setlength{\parindent}{0pt}
\setlength{\listparindent}{\parindent}
\setlength{\parsep}{0pt}
\setlength{\topsep}{3pt}}}{\end{list}}

\title{Beyond Retrieval: Improving Evidence Quality for LLM-based Multimodal Fact-Checking}

\author{Haoran Ou, 
  Gelei Deng,
  Xingshuo Han, 
  Jie Zhang,  
  Han Qiu, 
  Shangwei Guo, \\ 
  \textbf{Tianwei Zhang},  
  \textbf{Kwok-Yan Lam} 
  }

\begin{document}
\maketitle
\begin{abstract}

The increasing multimodal disinformation, where deceptive claims are reinforced through coordinated text and visual content, poses significant challenges to automated fact-checking. Recent efforts leverage Large Language Models (LLMs) for this task, capitalizing on their strong reasoning and multimodal understanding capabilities. Emerging retrieval-augmented frameworks further equip LLMs with access to open-domain external information, enabling evidence-based verification beyond their internal knowledge. Despite their promising gains, our empirical study reveals notable shortcomings in the external search coverage and evidence quality evaluation. To mitigate those limitations, we propose \Name, an end-to-end framework for automated multimodal fact-checking. It introduces a novel \textit{evidence retrieval strategy} that improves evidence coverage and filters useless information from open-domain sources, enabling the extraction of high-quality evidence for verification. Extensive experiments demonstrate that \Name achieves an accuracy of 88.3\% on two public multimodal disinformation datasets and 90.2\% on newly emerging claims. Compared with existing evidence retrieval strategies, our approach improves verification accuracy by up to 30.8\%, highlighting the critical role of evidence quality in LLM-based disinformation verification.

\end{abstract}

\noindent\emph{``Truth is Aletheia: the unconcealment of what is.''}

\begin{flushright}
\textasciitilde{} Martin Heidegger
\end{flushright}


\section{Introduction}
\label{sec:intro}

\begin{figure}[t]
    \centering
    \includegraphics[width=\linewidth]{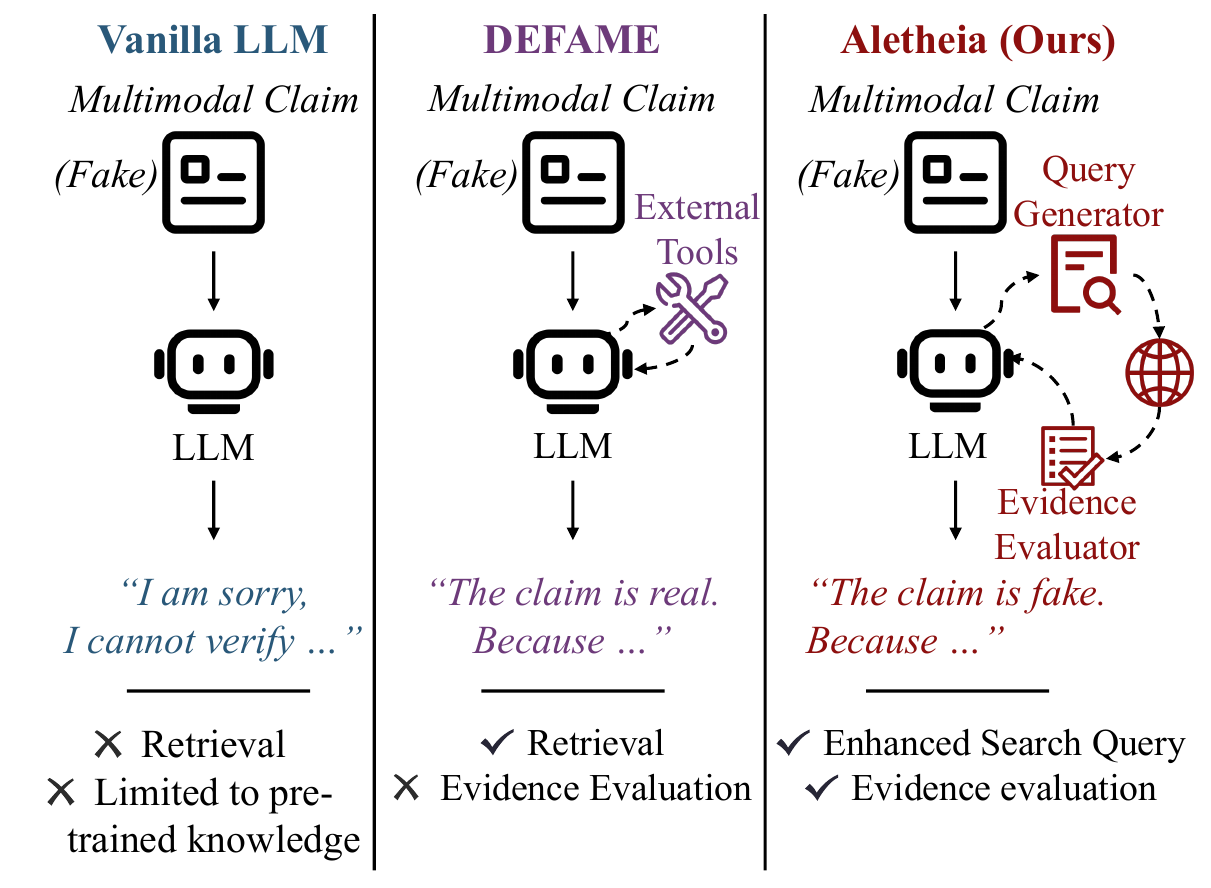}
    \caption{Comparisons of different strategies (including our \Name) for multimodal fact-checking.}
    \label{fig:fact_check}
\end{figure}

The proliferation of social media has led to a sharp growth in online content. It also makes disinformation, which is misleading or deceptive, become increasingly prevalent. Such content poses challenges for information reliability, thereby presenting considerable risks to real-world safety. Traditional disinformation detection solutions primarily focus on textual content and utilize deep learning techniques~\cite{misinformation-detection-rnn, misinformation-detection-cnn, misinformation-detection-attention}. With the increase in multimodal disinformation, recent studies leverage cross-modal feature alignment to assess consistency between text and images~\cite{related-work-multimodal-misinformation-detection-1, related-work-multimodal-misinformation-detection-2}.

The recent advances of Large Language Models (LLMs) bring new opportunities for disinformation detection due to their impressive reasoning~\cite{llm-reasoning-1,llm-reasoning-2} and multimodal processing~\cite{llm-multimodality} abilities. Researchers build detection systems atop LLMs~\cite{related-work-llm-1, related-work-llm-2, related-work-llm-3, related-work-llm-4}, improving their performance with techniques such as chain-of-thought prompting~\cite{llm-based_cot}, claim decomposition~\cite{llm-based_claim_decomposition}, question-guided prompting~\cite{llm-based_q_guided}, etc. However, the above standalone frameworks, including both DNN-based and LLM-based approaches, are subject to their training data cutoff(Figure~\ref{fig:fact_check}, left). As a result, they are constrained by their internal knowledge and struggle to effectively verify claims that fall outside their knowledge boundaries.

To mitigate this limitation, a promising strategy is to incorporate open-domain external information using third-party search tools~\cite{background-afc-2, defame, qi2024sniffer, wang2024mmidr, xuan2024lemma, tonglet2024image, du2023improving}. By providing external evidence, LLMs can more accurately recognize non-factual information compared to vanilla models. However, existing systems following this strategy exhibit a notable limitation: they largely rely on the internal safety and ranking mechanisms of the third-party tools, while lacking a rigorous or explicit strategy to assess the quality of the retrieved evidence (Figure~\ref{fig:fact_check}, middle). Therefore, the collected evidence could be potentially noisy, low-quality, or misleading, substantially affecting the detection accuracy. 

We conduct a targeted empirical study to validate the above argument (Section \ref{sec:insights}). We compare two evidence-assisted verification settings: directly providing LLMs with expert-written evidence (groundtruth) versus the state-of-the-art agent-based retrieval framework DEFAME \cite{defame}. The experiments disclose two key factors contributing to DEFAME's verification failures: (i) limited coverage in evidence retrieval, and (ii) the inclusion of noisy or weakly relevant evidence that undermines reliable verification.  
These observations highlight that evidence quality plays a decisive role in verification accuracy.

Building on these findings, we design \Name, an end-to-end framework for multimodal disinformation verification. Compared to existing solutions, the core component of \Name is a novel \textbf{evidence retrieval strategy} (Figure~\ref{fig:fact_check}, right). To improve evidence coverage, \Name begins with retrieval-oriented multimodal claim interpretation. It generates a set of structured sub-claims used for search, which capture distinct factual aspects that require verification. Rather than directly using the original claim or sub-claims produced by generic decomposition strategies as search inputs, \Name performs principled search query reformulation at the input level to achieve broader and more targeted evidence coverage. To reduce noise in retrieved results, \Name applies a structured evidence evaluation pipeline. First, it filters out evidence from unreliable sources. It then scores the remaining candidates based on their relevance to the claim and completeness of the information they provide. The top-ranking candidates are finally selected as high-quality evidence for verification.

We evaluate the effectiveness of \Name through extensive experiments on two public multimodal disinformation datasets~\cite{related-work-afc-2, related-work-afc-3}. \Name achieves an accuracy of 88.3\% and demonstrates stronger generalization compared to four deep learning-based baselines~\cite{related-work-afc-2, related-work-afc-3, spotfakeplus, pre_cofactv2}. We further assess the practical efficiency of \Name using a self-constructed dataset that reflects newly emerging claims. \Name attains a 90.2\% success rate in automatic evidence retrieval and claim verification, with an average cost of 0.11 USD and latency of 24.6 seconds per claim. Compared to the state-of-the-art agent system DEFAME~\cite{defame}, \Name is more accurate, efficient, and cost-effective. We attribute the improvement to the claim interpretation for broader retrieval coverage and evidence evaluation for filtering noisy or insufficient evidence. Additional ablation studies and retrieval efficiency analyses further validate the effectiveness of these two components in improving verification accuracy and robustness.

\section{Related Work}
\label{sec:background}

\subsection{Direct Disinformation Detection}

Early work on disinformation detection focuses on identifying false content directly from textual features using supervised learning models~\cite{related-work-text-misinformation-detection-1,related-work-text-misinformation-detection-2,related-work-text-misinformation-detection-3}. Recent studies extend to multimodality by jointly combining textual and visual signals through feature fusion or cross-modal alignment~\cite{related-work-multimodal-misinformation-detection-1,related-work-multimodal-misinformation-detection-2,related-work-multimodal-misinformation-detection-3}.

The rising of large language models (LLMs) have opened new opportunities for disinformation detection and fact-checking due to their strong reasoning and generation capabilities. Prior studies explore the use of LLMs for disinformation detection, showing promising results but still lagging behind human fact-checkers~\cite{related-work-llm-1,related-work-llm-2}. To improve performance, several works~\cite{related-work-llm-3,related-work-llm-4} propose prompting strategies or structured reasoning frameworks such as chain-of-thought prompting~\cite{llm-based_cot}, claim decomposition~\cite{llm-based_claim_decomposition}, question-guided prompting~\cite{llm-based_q_guided}, etc.

While these approaches achieve strong performance on benchmarks, their reliance on fixed training data often leads to poor generalization when encountering novel topics, writing styles, or emerging events. Another limitation is a lack of interpretability. Although LLM-based methods improve transferability to some extent, they still struggle to verify claims that fall outside their training data cutoff due to inherent knowledge limitations.

\subsection{Evidence-Driven Automated Fact Check}

Automated fact-checking (AFC), which verifies claims by retrieving and reasoning over external evidence, can effectively mitigate the above problems. A typical AFC framework consists of claim analysis, evidence retrieval, and verdict prediction with justification~\cite{background-afc-1, background-afc-5}. Compared to direct detection methods, AFC systems generally achieve higher robustness by grounding decisions in supporting evidence~\cite{background-afc-6}. 

A central challenge in AFC lies in evidence retrieval. Prior approaches retrieve evidence from curated corpora such as Wikipedia or fact-checking archives~\cite{background-afc-3,related-work-retrieval-1-2}, While such static sources provide high-quality information, they are inherently limited in coverage and timeliness, making them insufficient for verifying emerging claims or breaking news. Moreover, maintaining and updating curated corpora requires substantial manual effort and time. To address these limitations, recent research retrieves evidence dynamically from open-domain web sources. The retrieved evidence is subsequently incorporated into DNN-based~\cite{related-work-afc-1,related-work-afc-2} or LLM-based~\cite{background-afc-2, defame, qi2024sniffer, wang2024mmidr, xuan2024lemma, tonglet2024image, du2023improving} verification models to support claim verification.

These methods demonstrate that retrieved evidence can improve verification performance compared to standalone models, as it mitigates the problem that base models are constrained by their internal knowledge boundaries. However, there is a potential risk that they deeply trust the search results returned by third-party tools. Therefore, many existing approaches either lack explicit strategies or rely on limited and naive mechanisms to assess the quality and source credibility of retrieved evidence.

\section{Motivation}
\label{sec:insights}

We conduct a targeted empirical study for the failure analysis of existing fact-checking frameworks, highlighting the importance of evidence quality. We choose DEFAME \cite{defame}, the state-of-the-art agent system that automatically searches evidence for misinformation verification. We compare it with the ground truth, where LLMs are directly provided with evidence written by human experts. 
We construct an evaluation dataset by collecting 226 multimodal claims from Reuters, specifically designed to assess LLMs' performance in real-world disinformation verification. To ensure fairness, all the samples are published after the LLMs' knowledge cutoff date. Table~\ref{tab:error_breakdown} shows the overall detection accuracy, and the evidence error breakdown. More experiment details are shown in Appendix~\ref{subsec:verification_with_evidence}.

\begin{table}[t]
    \centering
    
    \caption{Failure case analysis and breakdown.
    Percentages are computed over incorrect predictions only.
    \textbf{NEI} denotes that insufficient information is retrieved.
    \textbf{NSY} denotes that evidence is misleading or low-quality.
    \textbf{OTH} include intrinsic LLM errors or others. \textbf{ACC} is the detection accuracy.}
    \label{tab:error_breakdown}
    \resizebox{\linewidth}{!}{
    \begin{tabular}{lcccc}
    \hline
    \textbf{Evidence Source} & \textbf{NEI (\%)} & \textbf{NSY (\%)} & \textbf{OTH (\%)} & \textbf{ACC (\%)}\\
    \hline
    Human-written  & 0.0  & 0.0  & 100.0 & 90.3\\
    DEFAME     & 48.8 & 46.3  & 4.9 & 63.7 \\
    \hline
    \end{tabular}}
\end{table}

\begin{figure*}[t]
    \centering
    \includegraphics[width=0.9\linewidth]{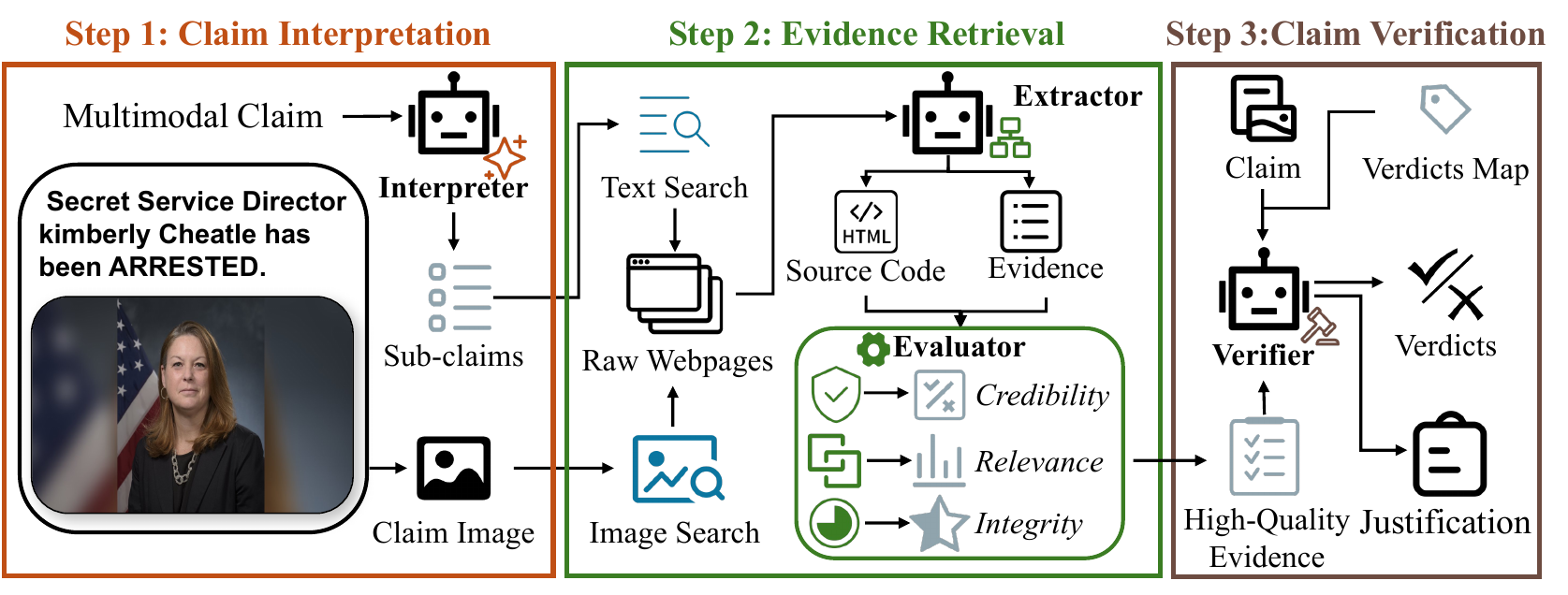}
    \caption{
    Overview of \Name, improving evidence quality for multimodal fact-checking by (1) retrieval-oriented multimodal claim interpretation and (2) structured evidence quality evaluation.}
    \label{fig:framework}
\end{figure*}

First, compared to the ground truth where LLMs are provided with evidence from human experts (90.3\%), a clear performance gap remains when adopting evidence from automated retrieval of DEFAME (63.7\%). 
When analyzing the failure cases of DEFAME, 48.8\% of errors comes from that the automated retrieval component does not collect sufficient information or any relevant sources to verify the claim, indicated by the LLM's responses. 
Excluding uncontrolled factors, such as the limited search capabilities of third-party tools or the scarcity of relevant information on the web, we attribute this failure primarily to the shortages of the generated search queries. 
Specifically, the queries are inaccurate or fail to adequately cover the key factual factors of the multimodal claim, making the framework fail to retrieve sufficient evidence to support a success verification.

\begin{tcolorbox}[colback=blue!5!white,colframe=gray!75!black,left=1mm, right=1mm, top=0.5mm, bottom=0.5mm, arc=1mm]
\textbf{Insight 1:} Incomplete evidence retrieval leaves LLM-based frameworks unable to determine whether a claim is true or false. 
\end{tcolorbox}

Second, another failure pattern (46.3\%) observed in DEFAME is the misleading evidence, where verification outcomes are flipped (e.g., predicting false claims as true or true claims as false). 
Under the same verification conditions, LLMs are able to produce correct verification results when provided with high-quality evidence from human. 
We therefore attribute this type of failure to the presence of noisy or weakly relevant evidence retrieved by DEFAME, which can distort LLM reasoning and lead to incorrect judgments.

\begin{tcolorbox}[colback=blue!5!white,colframe=gray!75!black,left=1mm, right=1mm, top=0.5mm, bottom=0.5mm, arc=1mm]
\textbf{Insight 2:} Noisy or weakly relevant evidence can actively mislead the verification process. 
\end{tcolorbox}

Taken together, these findings indicate that evidence quality and coverage play a decisive role in achieving reliable and accurate fact-checking.

\section{\Name}\label{sec:methodology}


Inspired by the above findings, we propose \Name, an end-to-end framework for training-free, zero-shot multimodal fact-checking. The core of \Name is a novel evidence retrieval component, which optimizes the search scope and reliably evaluates information from the public web to support verification. Figure~\ref{fig:framework} provides an overview of \Name, consisting of three stages: \textit{Claim interpretation}, \textit{Evidence retrieval}, and \textit{Verification \& justification}. We detail each component in the following sections.

\subsection{Multimodal Claim Interpretation}
\label{subsec:claim_comprehension}

This stage aims to interpret multimodal claims and decompose them into retrieval-oriented sub-claims that facilitate effective evidence collection. Given both textual and visual inputs, \Name leverages an LLM to understand stated factual claims across modalities. It then generates structured sub-claims, each focusing on a specific fact that needs to be verified. Compared to directly using the original claim for search, these sub-claims provide more specific and diverse search queries. As a result, this enables the retrieval of broader and more relevant information. In addition to multimodal claim interpretation, \Name performs image-based analysis to support visual evidence retrieval and source tracing.

\subsection{Evidence Retrieval}
\label{subsec:evidence_retrieval}

\subsubsection{Locating Candidate Evidence Sources}

\Name leverages sub-claims generated above as search queries to retrieve information from the public web that is semantically or contextually related. 
Additional image-based retrieval is performed to supplement visual information that may not be fully represented in textual sub-claims. It can trace their origins or identify visually similar content, thereby detecting out-of-contextualization or misuse. This mixed retrieval method takes full advantage of multimodal information, thus enabling \Name to construct a pool of comprehensive candidate evidence sources.


\subsubsection{Content Extraction}
\label{subsubsec:content_extraction}

Candidate evidence sources retrieved from the web often contain noisy and heterogeneous content. Therefore, it is unsuitable to treat entire webpages as evidence for direct verification. To enable reliable fact-checking, \Name transforms raw web pages into structured evidence representations that capture the essential factual information required for claim verification.

Specifically, \Name formalizes raw web content into structured evidence consisting of eight factual dimensions: \textbf{People}, \textbf{Event}, \textbf{Location}, \textbf{Time}, \textbf{Reason}, \textbf{Background}, \textbf{Impact}, and \textbf{Follow-up}. These dimensions extend the criteria used in~\cite{fake-news-detection-policy}. \Name makes evidence more explicit and easier to operate on in this way, thereby improving the effectiveness of verification. Moreover, the proposed formulation provides a principled basis for evidence integrity evaluation in the following evidence evaluation task.

To complete this task, \Name leverages an LLM to extract the relevant factual content from each candidate source. Given the textual content of a webpage, the LLM is guided to extract information corresponding to the defined evidence dimensions. During this process, non-informative webpage elements such as headers, footers, advertisements, and other boilerplate content are filtered out, while essential factual information is preserved. Implementation details are provided in Appendix~\ref{sec:prompt_temp}.

\subsubsection{Evaluating Evidence}
\label{subsubsec:evaluating_evidence}

Not all retrieved evidence is suitable for claim verification. To ensure reliability, \Name evaluates candidate evidence along three complementary dimensions: \textit{credibility}, \textit{relevance}, and \textit{integrity}. These criteria assess whether evidence comes from a trustworthy source, is semantically aligned with the claim, and provides sufficient factual information for verification.

Formally, given a claim with semantic representation $C$ and an extracted evidence set $E=\{(e_i, link_i)\}$, where $e_i$ denotes the evidence text and $link_i$ its source URL, \Name assesses the quality of each evidence item as described below.

\begin{algorithm}[t]
\caption{Evidence Quality Evaluation}
\label{al:evidence-evaluation-algorithm}
\KwIn{Claim representation $C$; candidate evidence set $E=\{(link_i, e_i)\}_{i=1}^n$}
\KwOut{Ranked evidence set $\hat{E}$}

\BlankLine
\textbf{Stage 1: Credibility filtering.} \\
$\mathcal{I} \leftarrow \textsc{FilterByCredibility}(E)$ \tcp*{source-level reliability check}

\BlankLine
\textbf{Stage 2: Evidence scoring.} \\
\ForEach{$(link_i, e_i) \in \mathcal{I}$}{
    $r_i \leftarrow \textsc{Relevance}(C, e_i)$\;
    $m_i \leftarrow \textsc{Integrity}(e_i)$\;
    $q_i \leftarrow \alpha \cdot r_i + (1-\alpha) \cdot m_i$\;
}

\BlankLine
\textbf{Stage 3: Ranking.} \\
$\hat{E} \leftarrow \textsc{RankByScore}(\{(e_i, q_i)\})$\;

\Return $\hat{E}$
\end{algorithm}

\begin{packeditemize}

    \item \textbf{Credibility.}
    It evaluates whether an evidence source is trustworthy. \Name performs credibility assessment at the source level, as unreliable sources can undermine verification. Specifically, \Name first filters out evidence from low-credibility or biased websites using publicly available blacklists~\cite{wikipedia_blacklist, media_bias}. For the remaining sources, \Name applies an automated credibility assessment model~\cite{credibility_model} that predicts webpage reliability by leveraging multi-dimensional features, including content quality, page structure, and link-based features. Only evidence from sources that meet a predefined credibility threshold is retained.

    \item \textbf{Relevance.}
    It measures the semantic alignment between the claim and the evidence content. \Name encodes the multimodal claim using BLIP-2~\cite{blip2} and computes a relevance score by comparing the semantic representation of the claim $C$ with the evidence text $e_i$ using cosine similarity. A higher relevance score indicates that the evidence is more closely related to the factual content of the claim.

    \item \textbf{Integrity.}
    It evaluates whether the evidence provides sufficiently complete factual information for verification. \Name uses ChatIE~\cite{chatie} to extract structured event arguments from each evidence item. Each argument consists of a predefined role (e.g., \emph{Person}) and its corresponding textual content (e.g., a specific film actor's name). The coverage of the extracted roles is aligned with the structured evidence schema introduced in Section~\ref{subsubsec:content_extraction}. Integrity is then measured as the proportion of roles whose corresponding content is successfully extracted among all predefined roles.
    
\end{packeditemize}

The evidence quality evaluation procedure is summarized in Algorithm~\ref{al:evidence-evaluation-algorithm}. 
Given a claim representation $C$ and a set of candidate evidence items $E$, \Name first filters out evidence from unreliable sources based on credibility assessment. 
For the remaining candidates, it further evaluates evidence quality by jointly considering semantic relevance to the original claim and factual integrity. 
These two scores are combined using a weighted aggregation scheme to produce a final evidence quality score, which is used to rank evidence candidates for subsequent verification.
The weight $\alpha$ settings are provided in Appendix~\ref{subsec:parameters_justification}.


\subsection{Claim Verification}
\label{subsec:claim-verification}





\Name formulates the claim verification task as a binary classification problem. Although professional fact-checking organizations often adopt fine-grained verdict labels (e.g., \emph{Mostly True}, \emph{Partially False}), such labels introduce subjective and ambiguous decision boundaries, as labeling standards can vary across different organizations. The binary formulation provides a clearer and more reliable decision criterion for automated verification. Consequently, \Name maps all fine-grained verdicts produced by fact-checking agents to \emph{true} or \emph{false}. Details are provided in Appendix~\ref{sec:detailed_settings}, Table~\ref{tab:complete_labels_map}. This design improves verification accuracy and robustness by reducing label ambiguity while retaining the primary goal of assessing claim truthfulness.

Specifically, \Name guides an LLM to verify the truthfulness of the claim and generate a justification grounded in the retrieved evidence. To improve reasoning stability and reduce interference from long inputs, \Name adopts a structured, stage-wise interaction protocol that separates task initialization, evidence incorporation, and verification. Finally, the structured justification for its verdict explains how it supports or refutes the claim. This interpretable verification process enhances the transparency, thereby strengthening the credibility of the verdict. All prompt templates and output formats used in this phase are provided in Appendix~\ref{sec:prompt_temp}. We provide a concrete illustrative example in Appendix~\ref{sec:illustrative_example} to demonstrate how \Name operates in a real-world multimodal fact-checking scenario.

\section{Evaluation}\label{sec:evaluation}

We evaluate the effectiveness of \Name under different verification settings.

\begin{itemize}[leftmargin=*,itemsep=1pt,topsep=0pt,parsep=1pt]

    \item \textbf{RQ1 (Benchmark Evaluation)} How effective is \Name in verifying multimodal disinformation on public benchmark datasets?

    \item \textbf{RQ2 (Open-World Verification)} Can \Name verify disinformation in an open-world setting by automatically retrieving evidence?

    \item \textbf{RQ3 (Ablation Study)} How does each component of the proposed framework contribute to overall verification performance?
    
\end{itemize}

\subsection{Experimental Setup}
\label{subsec:Experimental_Setup}

\noindent\textbf{Datasets.}
We evaluate \Name under different verification settings corresponding to RQ1--RQ3.
For \textbf{RQ1}, we adopt two public multimodal disinformation benchmarks, Mocheg~\cite{related-work-afc-3} and MR2~\cite{related-work-afc-2}, which have been widely used in prior work.
These datasets provide multimodal claims with ground-truth labels and supporting evidence. For \textbf{RQ2} and \textbf{RQ3}, we construct a new dataset, \textit{MMDV} (Multi-Source Multimodal Disinformation Verification Dataset). Unlike existing benchmarks, MMDV contains only multimodal claims and labels, without any predefined supporting evidence. Moreover, all claims in MMDV are published after the knowledge cutoff dates of the evaluated LLMs, ensuring a fair assessment that prevents reliance on memorized knowledge. This better reflects open-world verification scenarios.
Detailed dataset statistics and construction procedures are provided in Appendix~\ref{subsec:settings}.

\begin{table}
    \caption{
    Comparison of baseline methods across key capabilities.
    \textbf{Multimodal} denotes the support for multimodal claim verification;
    \textbf{Web Search} denotes the ability to retrieve information from the public web;
    \textbf{Evidence Evaluation} denotes explicit assessment of evidence quality;
    \textbf{Explainability} denotes the ability to generate interpretable justifications.
    }
    \centering
    \resizebox{\linewidth}{!}{
    \begin{tabular}{ccccc}
    \toprule
         & \multirow{2}{*}{Multimodal} & Web & Evidence & \multirow{2}{*}{Explainability} \\ 
         & & Search & Evaluation & \\
         \midrule
    Pre-CoFactv2     & \ding{51} & \ding{55} & \ding{55} & \ding{55} \\
    End2End     &  \ding{51} & \ding{55} & \ding{55} & \ding{51} \\
    RB    & \ding{51} & \ding{51} & \ding{55} & \ding{55} \\
    SpotFakePlus  & \ding{51} & \ding{55} & \ding{55} & \ding{55} \\ 
    DEFAME & \ding{51} & \ding{51} & \ding{55} & \ding{51} \\
    \midrule
    \Name & \ding{51} & \ding{51} & \ding{51} & \ding{51} \\
    \bottomrule

    \end{tabular}}
    \label{tab:baselines}
\end{table}

\begin{table*}[t]
  \caption{Verification performance on the Mocheg and MR2 benchmark datasets.}
  \centering
  \resizebox{\linewidth}{!}{
  \begin{tabular}{c||cccc||cccc}
    \hline
    \multirow{2}{*}{} & \multicolumn{4}{c||}{Mocheg} & \multicolumn{4}{c}{MR2} \\
    & Accuracy & Precision & Recall & F1 & Accuracy & Precision & Recall & F1 \\

    \hline
    Pre-CoFactv2    & 46.7\% & 51.1\% & 46.3\% & 41.1\% & 60.2\% & 64.0\% & 57.5\% & 57.2\%\\
    End2End      & 54.5\% & 55.8\% & 54.2\% & 51.7\% & 54.4\% & 55.9\% & 54.1\% & 51.8\% \\
    RB           & 37.5\% & 44.6\% & 37.0\% & 25.6\% & 62.8\% & 66.6\% & 60.3\%  & 59.2\% \\
    SpotFakePlus & 53.0\% & 54.8\% & 54.7\% &  52.9\% & 54.0\% & 55.6\% & 54.7\% & 52.2\% \\
    DEFAME & 60.1\% & 59.7\% & 61.2\% & 60.4\% & 70.2\% & 71.1\% & 70.6\% & 70.8\%\\
    \Name (Llama 3.2-vision) & 61.1\% & 62.4\% & 61.1\% & 60.0\% & 34.9\% & 17.4\% & 50.0\% & 25.9\%\\
    \Name (Qwen-vision) & 47.3\% & 40.3\% & 47.4\% & 34.6\% & 66.2\% & 34.6\% & 46.9\% & 39.8\%\\
    \Name (Gemini-1.5-flash) & 64.9\% & 65.1\% & 64.9\% & 65.0\% & 73.8\% & 63.1\% & 74.7\% & 68.4\% \\
    \Name (GPT-4o) & \textbf{73.8\%} & \textbf{75.9\%} & \textbf{73.9\%} & \textbf{73.2\%} & \textbf{88.3\%} & \textbf{88.8\%} & \textbf{88.2\%} & \textbf{88.3\%} \\
    \hline
  \end{tabular}}

  \label{tab:performance}
\end{table*}

\begin{table*}[t]
    \caption{
    Transferability performance on the Mocheg and MR2 benchmark datasets.
    Models are trained on one dataset and evaluated on the other.
    $\downarrow$ indicates performance degradation relative to Table~\ref{tab:performance}.
    }

  \centering
  \resizebox{\linewidth}{!}{
  \begin{tabular}{c||cccc||cccc}
    \hline
    \multirow{2}{*}{} & \multicolumn{4}{c||}{Mocheg(MR2)} & \multicolumn{4}{c}{MR2(Mocheg)} \\
    & Accuracy & Precision & Recall & F1 & Accuracy & Precision & Recall & F1 \\

    \hline
    Pre-CoFactv2    & 34.2\% $\downarrow$ & 32.8\% $\downarrow$ & 33.8\% $\downarrow$ & 23.3\% $\downarrow$ & 34.6\% $\downarrow$ & 36.0\% $\downarrow$ & 36.8\% $\downarrow$ & 32.2\% $\downarrow$\\
    End2End      & 36.2\% $\downarrow$ & 38.6\% $\downarrow$ & 38.7\% $\downarrow$ & 29.2\% $\downarrow$ &  35.5\% $\downarrow$ & 37.6\% $\downarrow$ & 37.9\% $\downarrow$ & 28.8\% $\downarrow$ \\
    RB           & 33.5\% $\downarrow$ & 29.3\% $\downarrow$ & 33.1\% $\downarrow$ & 19.2\% $\downarrow$ & 34.2\% $\downarrow$ & 41.9\% $\downarrow$ & 36.2\% $\downarrow$ & 29.0\% $\downarrow$ \\
    SpotFakePlus & 34.3\% $\downarrow$ & 29.6\% $\downarrow$ & 34.8\% $\downarrow$ & 20.9\% $\downarrow$ & 36.5\% $\downarrow$ & 23.8\% $\downarrow$ & 36.4\% $\downarrow$ & 28.1\% $\downarrow$ \\
    DEFAME & 60.1\% & 59.7\% & 61.2\% & 60.4\% & 70.2\% & 71.1\% & 70.6\% & 70.8\%\\
    \Name (Llama 3.2-vision) & 61.1\% & 62.4\% & 61.1\% & 60.0\% & 34.9\% & 17.4\% & 50.0\% & 25.9\%\\
    \Name (Qwen-vision) & 47.3\% & 40.3\% & 47.4\% & 34.6\% & 66.2\% & 34.6\% & 46.9\% & 39.8\%\\
    \Name (Gemini-1.5-flash) & 64.9\% & 65.1\% & 64.9\% & 65.0\% & 73.8\% & 63.1\% & 74.7\% & 68.4\% \\
    \Name (GPT-4o) & \textbf{73.8\%} & \textbf{75.9\%} & \textbf{73.9\%} & \textbf{73.2\%} & \textbf{88.3\%} & \textbf{88.8\%} & \textbf{88.2\%} & \textbf{88.3\%} \\
    \hline
  \end{tabular}}
  \label{tab:generalization}
\end{table*}

\noindent\textbf{Baselines.}
We evaluate \Name with four multimodal LLM backbones, including two commercial models (GPT-4o and Gemini-1.5-Flash) and two open-source alternatives (Llama-3.2-Vision-11B~\cite{llama-3.2-vision} and Qwen-Vision-7B~\cite{qwen-vision}).
We compare \Name against five representative multimodal disinformation verification baselines: End2End~\cite{related-work-afc-3}, RB~\cite{related-work-afc-2}, Pre-CoFactv2~\cite{pre_cofactv2}, SpotFakePlus~\cite{spotfakeplus}, and DEFAME~\cite{defame}. 
Table~\ref{tab:baselines} summarizes baseline capabilities along four dimensions that are essential for multimodal fact-checking: multimodal processing, web search, evidence evaluation, and explainability.
Existing methods cover some of these aspects, but typically lack explicit evidence quality assessment or structured justification generation. \Name integrates all four capabilities to support end-to-end and interpretable fact-checking.
Details are provided in Appendix~\ref{subsec:settings}.

It is worth noting that OpenAI and Google have integrated online search functionality into their LLMs~\cite{gpt_web_search, gemini_search_grounding}, allowing models to retrieve relevant online information. Although not explicitly designed for fact-checking, they have the potential for evidence-based verification. We evaluate the performance of these search-enhanced models on fact-checking tasks in Appendix~\ref{sec:llm-search}. Our analysis reveals several limitations that reduce their effectiveness for multimodal disinformation detection. Specifically, these models are limited to the single textual modality: they neither support explicit visual understanding nor enable image-based retrieval. Consequently, their performance degrades substantially on two types of claims: (1) claims conveyed solely through images without accompanying textual descriptions, and (2) multimodal claims containing both text and images, where the key factual information is expressed predominantly through visual content. These limitations indicate that current search-enhanced LLMs are not well-suited for multimodal fact-checking tasks.

\noindent\textbf{Settings.}
We evaluate all methods using standard metrics, including Accuracy, Precision, Recall, and F1-score.
We implement four variants of \Name by instantiating GPT-4o, Gemini-1.5-Flash, Llama-3.2-Vision-11B, and Qwen-Vision-7B as both the evidence extractor and verifier.
For each variant, the same LLM is used consistently across components, with temperature set to zero to ensure deterministic outputs.
Additional implementation and environment details are provided in Appendix~\ref{subsec:settings}.

\subsection{(RQ1) Benchmark Evaluation}

We evaluate \Name on two public benchmarks, Mocheg and MR2, under standard and cross-dataset transferability settings.

\noindent\textbf{Benchmark Performance.}
Table~\ref{tab:performance} reports the verification performance on each dataset.
Overall, \Name consistently outperforms all baseline methods on both benchmarks, except when instantiated with LLaMA-3.2-Vision.
Among different backbones, \Name achieves stronger performance with commercial LLMs than open-source models.
In particular, \Name with GPT-4o achieves the best results on both datasets, reaching 73.8\% accuracy on Mocheg and 88.3\% on MR2.

\noindent\textbf{Transferability.}
To assess robustness across domains, we evaluate models trained on one dataset and tested on the other.
As shown in Table~\ref{tab:generalization}, \Name maintains strong performance across datasets without retraining, whereas training-based baselines suffer substantial performance degradation.
For example, Pre-CoFactv2 and RB drop to near-random performance when evaluated on unseen datasets.
DEFAME, which is also training-free and LLM-based, exhibits relatively stable performance across datasets.
However, its overall accuracy remains consistently lower than that of \Name.
This suggests that while training-free designs help mitigate domain shift, effective evidence retrieval and evaluation are critical for achieving robust verification performance.

\subsection{(RQ2) Open-World Verification}
\label{subsec:verification_with_evidence}

We evaluate the performance of \Name in an open-world verification setting, where no supporting evidence is provided, and models must autonomously retrieve information from the web.
Therefore, in this setting, only methods with web search capability (RB, DEFAME, and \Name) can leverage external evidence. The remaining baselines rely solely on the claim content.

Table~\ref{tab:real-time_performance} reports the results.
Overall, evidence-based methods substantially outperform content-only approaches, highlighting the critical role of external evidence and its quality in open-world verification.
Baselines without retrieval capability perform poorly, with accuracy even less than random guessing.
RB, despite supporting web search, achieves the lowest accuracy, indicating limited robustness. DEFAME achieves competitive performance with an accuracy of 75.2\%, whose backbone model is GPT-4o. 
Although DEFAME slightly outperforms \Name instantiated with open-source LLMs, when using the same GPT-4o backbone, \Name consistently achieves higher accuracy, indicating the benefit of its evidence retrieval and evaluation design. In particular, \Name with GPT-4o achieves the highest accuracy of 90.2\%, followed by Gemini-1.5-Flash at 87.0\%.

These results demonstrate that while automatic retrieval is necessary for open-world verification, effective evidence selection and evaluation are also crucial for achieving high accuracy.
A detailed comparison of verification cost and running time is provided in Appendix~\ref{sec:cost}, showing that \Name is not only more accurate but also more efficient than existing alternatives, including DEFAME.

\begin{table}[t]
  \caption{Open-world verification performance on the MMDV dataset.}
  \centering
  \resizebox{\linewidth}{!}{
  \begin{tabular}{c||cccc}
    \hline
    & Accuracy & Precision & Recall & F1 \\

    \hline
    Pre-CoFactv2 & 41.9\% & 0.0\% & 0.0\% & 0.0\% \\
    End2End & 41.0\% & 20.5\% & 50.0\% & 29.1\% \\
    SpotFakePlus & 27.0\% & 17.9\% & 30.6\% & 21.8\% \\
    \hline
    RB    & 13.0\% & 16.4\% & 10.3\% & 12.7\% \\
    DEFAME & 75.2\% & 75.1\% & 75.7\% & 75.4\% \\
    \Name (Llama 3.2-vision) & 73.4\% & 70.8\% & 56.0\% & 48.1\% \\
    \Name (Qwen-vision) & 71.1\% & 39.6\% & 43.4\% & 41.6\% \\
    \Name (Gemini-1.5-flash) & 87.0\% & 87.8\% & 86.7\% & 86.8\% \\
    \Name (GPT-4o) & \textbf{90.2\%} & \textbf{89.9\%} & \textbf{90.0\%} & \textbf{89.8\%} \\
    \hline
  \end{tabular}}
  \label{tab:real-time_performance}
\end{table}

\subsection{(RQ3) Ablation Study}
\label{subsec:ablation_study}

\begin{table}[t]
  \caption{Ablation study on different components of \Name.}
  \centering
  \resizebox{\linewidth}{!}{
  \begin{tabular}{c||cccc}
    \hline
    & Accuracy & Precision & Recall & F1-score \\ 
    \hline
    w/o claim interpretation & 74.9\% & 73.2\% & 77.2\% & 75.1\% \\
    random evidence & 64.1\% & 72.2\% & 64.5\% & 60.8\% \\ 
    \hline
    Full system & \textbf{90.2\%} & \textbf{89.9\%} & \textbf{90.0\%} & \textbf{89.8\%}  \\
    \hline
  \end{tabular}}
  \label{tab:ablation_study}
\end{table}

We conduct an ablation study on the MMDV dataset to examine the contribution of key components in \Name. All variants in the study use the same underlying LLM, GPT-4o.
Results are summarized in Table~\ref{tab:ablation_study}.

\noindent\textbf{Multimodal Claim Interpretation.}
We remove the multimodal claim interpretation module and directly use the original claim text for retrieval. This results in a substantial performance drop, with accuracy decreasing from 90.2\% to 74.9\%. The result indicates that decomposing multimodal claims into retrieval-oriented sub-claims is critical for obtaining relevant evidence.

\noindent\textbf{Evidence evaluation.}
To assess the role of evidence evaluation, we replace the evidence evaluation module with random evidence sampling. This leads to a more pronounced performance degradation, reducing accuracy to 64.1\%. Under the same experimental setting, we further replace our evidence retrieval module with the RB method and observe a substantial drop in verification accuracy. The detailed experimental setup and results are reported in Section~\ref{sec:evidence-retrieval-comparison}. These results highlight the necessity of our evidence evaluation design.

\section{Conclusion}\label{sec:conclusion}

In this paper, we focus on the LLM-based multimodal fact-checking via evidence retrieval. We observe that existing solutions fall short in guaranteeing evidence coverage and quality. Driven by these limitations, we propose \Name, a pioneering automated fact-check framework to effectively detect multimodal disinformation. \Name integrates a novel evidence retrieval approach to acquire comprehensive, high-quality and relevant information from the public Internet, which can significantly improve the LLM's verification accuracy and rationality. Extensive experiments validate that \Name significantly outperforms state-of-the-art solutions over two multimodal benchmarks and a newly constructed dataset consisting of newly emerging claims. 

\section*{Limitations}

Despite the effectiveness of \Name in detecting disinformation across textual and image-based claims, it faces limitations when handling other modalities, such as audio or video. Detecting disinformation in these formats is especially challenging due to the complexity of temporal/visual-temporal signals, the need for synchronized multimodal reasoning, and the limited capabilities of current fact-checking frameworks in processing such content. Meanwhile, state-of-the-art tools like GPT-4o with web search or Gemini-1.5-flash with Google Search primarily support textual input and lack robust support for audio-visual analysis. This reveals a critical blind spot in the current research: the absence of reliable systems for verifying multimedia content, where key evidence is probably embedded in non-textual formats. These challenges suggest directions for integrating audio and video LLMs into \Name to support broader and more robust multimodal fact-checking. Moreover, \Name relies on evidence retrieved from publicly accessible web sources through search engines. The coverage of the underlying search engines can influence the effectiveness of verification. For newly emerging events, such as cases where authoritative information is primarily released through internal reports or institutional channels, relevant evidence may be limited or delayed in public search results. Overall, these limitations reflect practical constraints of open-world fact-checking and help delineate the scope in which \Name is most effective.

\section*{Ethical Considerations}

This work is conducted for research purposes and aims to support the assessment of disinformation by providing evidence-based analysis. \Name is intended to assist human judgment rather than authoritative certifications. This is the responsibility of dedicated fact-checking institutions. The framework operates solely on publicly accessible web content. No private, personal, or user-identifiable information is collected or processed in this work. While the analyzed disinformation may contain offensive content, it is used strictly for research analysis.
The manual annotation and analysis process is performed by the authors of this paper, who are domain experts of misinformation detection.
We also confirm that all artifacts used in this work (including datasets and models) are publicly available, and we use them in strict accordance with their respective licenses (e.g., MIT, Apache 2.0, or CC-BY) and intended use terms. Furthermore, the resources and code released in this work are intended to facilitate future research on AI security and robustness.




\bibliography{sample-base}

\appendix
\cleardoublepage
\appendix
\label{sec:appendix}

\section{Empirical Study}
\label{sec:study}

\begin{figure}
    \centering
    \includegraphics[width=1.0\linewidth]{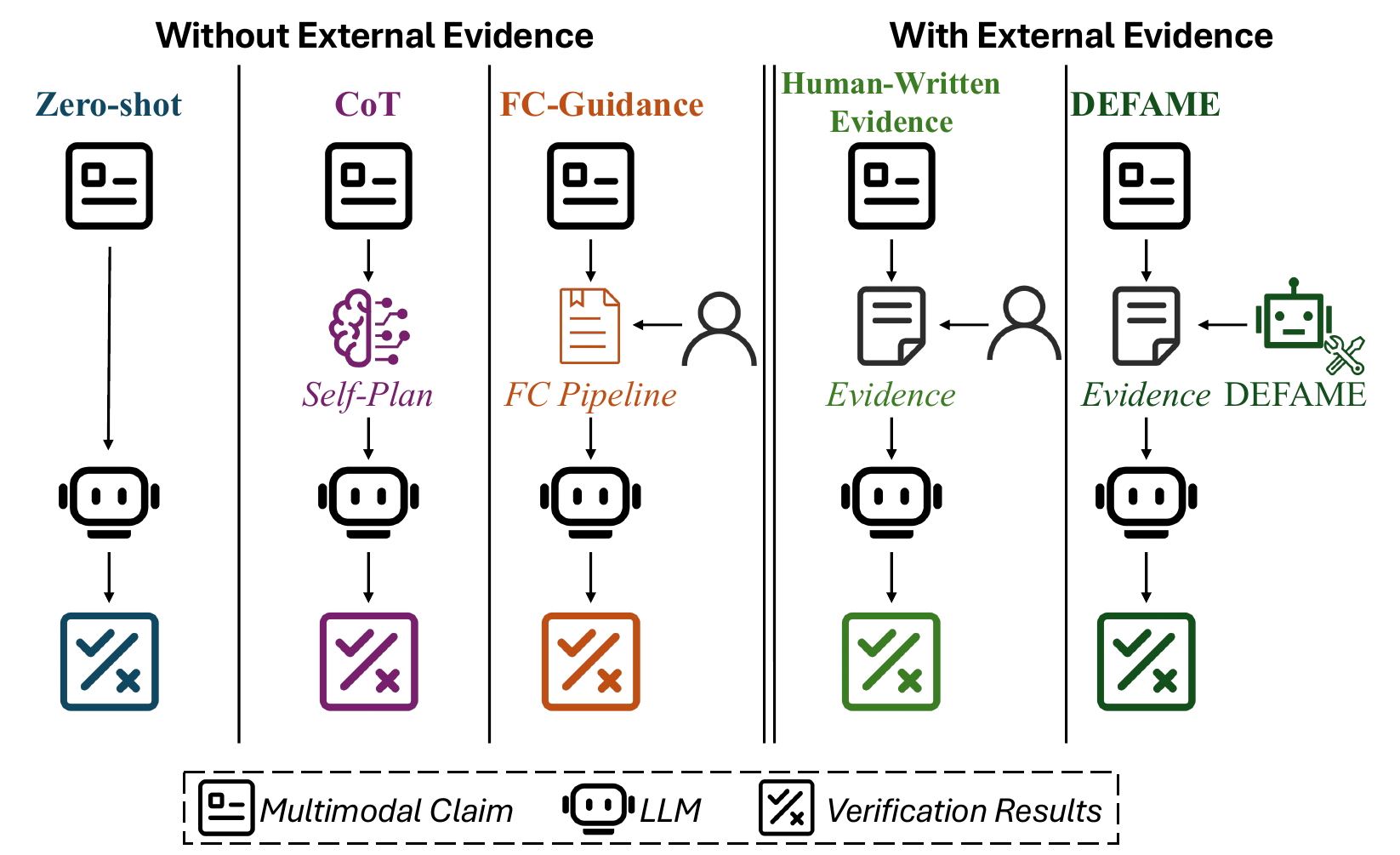} 
    \caption{LLM strategies for verifying disinformation.}
    \label{fig:empirical_study_overview}
\end{figure}

\begin{table}[t]
\caption{Experimental results of empirical study.}
\centering
\resizebox{\linewidth}{!}{
\begin{tabular}{ccccc}
    \toprule
    {\centering \textbf{Approach}}  & \multicolumn{2}{c}{\centering \textbf{Verification Rate}} & \multicolumn{2}{c}{\centering \textbf{Correctness Rate}} \\ 
    \cmidrule{2-5} & GPT-4o & Gemini-1.5-flash & GPT-4o & Gemini-1.5-flash \\ 
    \midrule
    Zero-shot  & 6.3\% & 4.5\% & 42.9\% & 40.2\% \\
    CoT        & 9.8\% & 7.4\%  & 43.2\% & 42.1\% \\
    FC guidance & 20.4\% & 91.2\% & 81.8\% & 78.6\% \\
    Human-written evidence & 100.0\% & 100.0\% & 90.3\% & 92.9\%\\
    DEFAME & 100.0\% & - & 63.7\% & -\\
    
\bottomrule

\end{tabular}}

\label{tab:empirical-study-1} 
\end{table}

To explore LLMs' behaviors in multimodal disinformation detection, we conduct a series of experiments to answer three research questions:

\begin{itemize}[leftmargin=*,itemsep=1pt,topsep=0pt,parsep=1pt]

\item\textbf{(RQ1)} How do standalone LLMs behave when directly verify the multimodal claims?

\item\textbf{(RQ2)} How do LLMs behave when verifying claims guided by the fact-checking pipeline?

\item\textbf{(RQ3)} How does external evidence quality affect LLMs' performance in verification?

\end{itemize}

We implement five distinct strategies to observe the behaviors of LLMs in verifying disinformation, as shown in Figure \ref{fig:empirical_study_overview}. These strategies are classified into two settings, depending on whether external evidence is provided in the verification process. First, we investigate how LLMs behave when verifying claims that fall outside their knowledge boundaries. More critically, we examine the performance of LLMs when provided with evidence from different sources.
We detail the evaluation below.

\subsection{Experimental Setup}

\subsubsection{Dataset Construction}
\label{subsubsec:rq1-dataset-construction}
We build a dataset for our evaluation, following two basic rules: (1)~\textit{Trustworthiness}: the dataset only contains verified news and disinformation as a valid benchmark. (2)~\textit{Timeliness}: the release date of the samples in the dataset is relatively recent and not included in the training set of the selected LLMs, so that we rule out potential biases arising from prior exposure to the disinformation.

To meet the trustworthiness requirement, we gather samples verified by reputable fact-checking agents. When disinformation appears on the Internet and raises significant public concern, authoritative entities such as government agencies respond promptly to combat it and maintain social stability. We consider samples that have undergone such fact-checking as validated and incorporate them into our dataset. Reuters\footnote{\url{https://www.reuters.com/}}, a popular news agency, serves as our primary data source due to its global reputation for impartiality, accuracy, and integrity in journalism. It has a ``news'' column that categorizes news into different sections according to their content. To ensure balance in the dataset, we collect an equal number of news sampled from various categories. We use the news caption as the claim to be verified and label it as true.
Additionally, Reuters has a ``fact-checking'' column that examines the disinformation circulating on social media (e.g., Twitter, Facebook). We collect these disinformation samples, including text and images, as negative examples. At the end of each fact-checking article, the authors provide a verdict, which includes labels such as false, satire, misleading, and others. We map all such labels to the false category. For the timeliness requirement, we review the release dates of the samples and filter out those that were published before the cutoff date of the LLM training sets\footnote{The knowledge cutoff dates of GPT-4o and Gemini-1.5-flash are October 2023 and November 2023, respectively.}. This ensures the selected samples are not in the knowledge base of the LLMs. 

\begin{table}[t]
    \caption{The statistics of the empirical study dataset. Reuters verdict labels are standardized into true/false.}
    \centering
    \resizebox{\linewidth}{!}{
    \begin{tabular}{ccccc}
    \toprule
    \textbf{Type} & \textbf{Count} & \textbf{Publish Date Range} & \textbf{Reuters Fact-check Verdicts} & \textbf{Standard Labels} \\ 
    \midrule
    News & 146 & Feb - Aug, 2024 & True   &  True \\ 
    \midrule
    Disinformation & 80 & Feb - Aug, 2024 & \makecell[c]{Misleading, Missing Context, \\ Altered, Synthetic Media, \\ Miscaptioned, Satire} &  False     \\
    \bottomrule

    \end{tabular}}
    \label{tab:reuters_labels_map}
\end{table}

The composition and property of the dataset are shown in Table~\ref{tab:reuters_labels_map}. Following those principles, we manually collect 226 samples from Reuters, and each sample consists of the claim and the corresponding label. There are 6 original verdicts of the disinformation and 1 verdict of the news. To balance the distribution of the disinformation and true news in the dataset~\cite{fever-dataset}, we select 146 positive examples and 80 negative ones. All of the samples are collected from articles released between February 2024 and August 2024, which are later than the knowledge cutoff dates of GPT-4o and Gemini-1.5-flash. 

For retrieval-based verification, we extend the dataset by supplementing each claim with additional evidence sourced from the same articles. To verify claims, Reuters journalists gather relevant content from authoritative sources and summarize it into evidence supporting or refuting the claim. We segment this evidence into paragraphs based on source and content, and collect corresponding justifications for further analysis.

\subsubsection{Evaluation Strategy}\label{subsubsec:rq1-evaluation-strategy}

We select two state-of-the-art LLMs that support the image modality: GPT-4o~\cite{gpt-4o} and Google Gemini-1.5-flash~\cite{gemini-1.5-flash}. We aim to observe how do these two models (1) verify the truthfulness of claims; (2) summarize logical reasons that support their verification.

For \textbf{RQ1}, the evaluation pipeline is shown in Figure~\ref{fig:empirical_study_overview}, "Zero-shot" and "CoT" column. We deploy two prompting techniques: zero-shot prompting~\cite{llm-reasoning-2} which directly provides the model with the task instruction and input without any task-specific examples, and Chain-of-Thoughts (CoT)~\cite{llm-reasoning-1} that guides the model to generate intermediate reasoning steps before generating the final answer. These two approaches have been widely applied to various tasks, such as question answering, text understanding, and mathematical reasoning~\cite{llm-reasoning-1, llm-reasoning-2}. The evaluation process begins by preparing the multimodal claim, which includes text and, if applicable, an associated image. For zero-shot prompting, the LLM is instructed to directly assess the claim's truthfulness using a standardized prompt: "\textit{Please verify the following claim. If you can verify the truthfulness of the claim, answer with `yes' and explain why it is true or false. If you cannot verify it, answer with `no' and provide the reason.}" The LLM's response is subsequently normalized for consistent analysis. For CoT, the LLM is guided to generate several logical steps to verify the disinformation, and is sent to the LLM for final verification along with the claim. 
For \textbf{RQ2}, the pipeline is shown in Figure~\ref{fig:empirical_study_overview}, "FC-Guidance" column. Instead of adopting the self-generated CoT, the models are prompted to follow the fact-checking (FC) pipeline to verify the claims in the following process: evidence retrieval, verdict prediction, and justification production.
For \textbf{RQ3}, the pipeline is shown in Figure~\ref{fig:empirical_study_overview}, "Human-Written Evidence" and "DEFAME" column. The former is a semi-automatic evaluation strategy: we first manually collect human-written evidence from the fact-check articles, which are expected to be highly relevant to the claims, then add the supporting evidence into the context, and submit the whole context to the LLMs for verification. For the latter, the evidence is automatically retrieved from the open-domain Internet by DEFAME.

For all methods, we analyze LLMs' responses in three aspects: (1) whether the LLM can perform the verification task; (2) what is the correctness rate if the LLMs can execute the task; (3) if there are any errors in the justification, e.g., hallucinations, that can affect the verification accuracy and reliability. This evaluation has two benefits. First, we can directly observe LLMs' performance in verification tasks under different settings; Second, manual evaluation explores what the challenge is if LLMs cannot verify the truthfulness of disinformation.

\subsubsection{Evaluation Metric}
We use the following metrics to benchmark the capability and detection accuracy of LLMs. 

\begin{packeditemize}

\item \textbf{Verification rate.} The proportion of claims for which the LLM attempts verification. If LLMs respond with ``\textit{Yes, I can do}'' or directly output the prediction results, we consider it a \textit{successful verification case}. Otherwise, outputs like ``\textit{No, I cannot do}'' denote that LLMs cannot complete the task. It is counted as a \textit{failed verification case}. This metric measures the percentage of \textit{successful verification cases} over the total claims.

\item \textbf{Correctness rate.} Among the \textit{successful verification cases}, we calculate the proportion of \textit{correctness verification cases} where the LLM prediction matches the ground truth.

\end{packeditemize}

\subsection{(RQ1) Standalone Verification}

We first explore how existing LLMs verify the truthfulness of claims solely based on claim content, without access to external information, and examine their specific behaviors when verification performance is poor.

\begin{figure*}[t]
    \centering
    \includegraphics[width=1.0\linewidth]{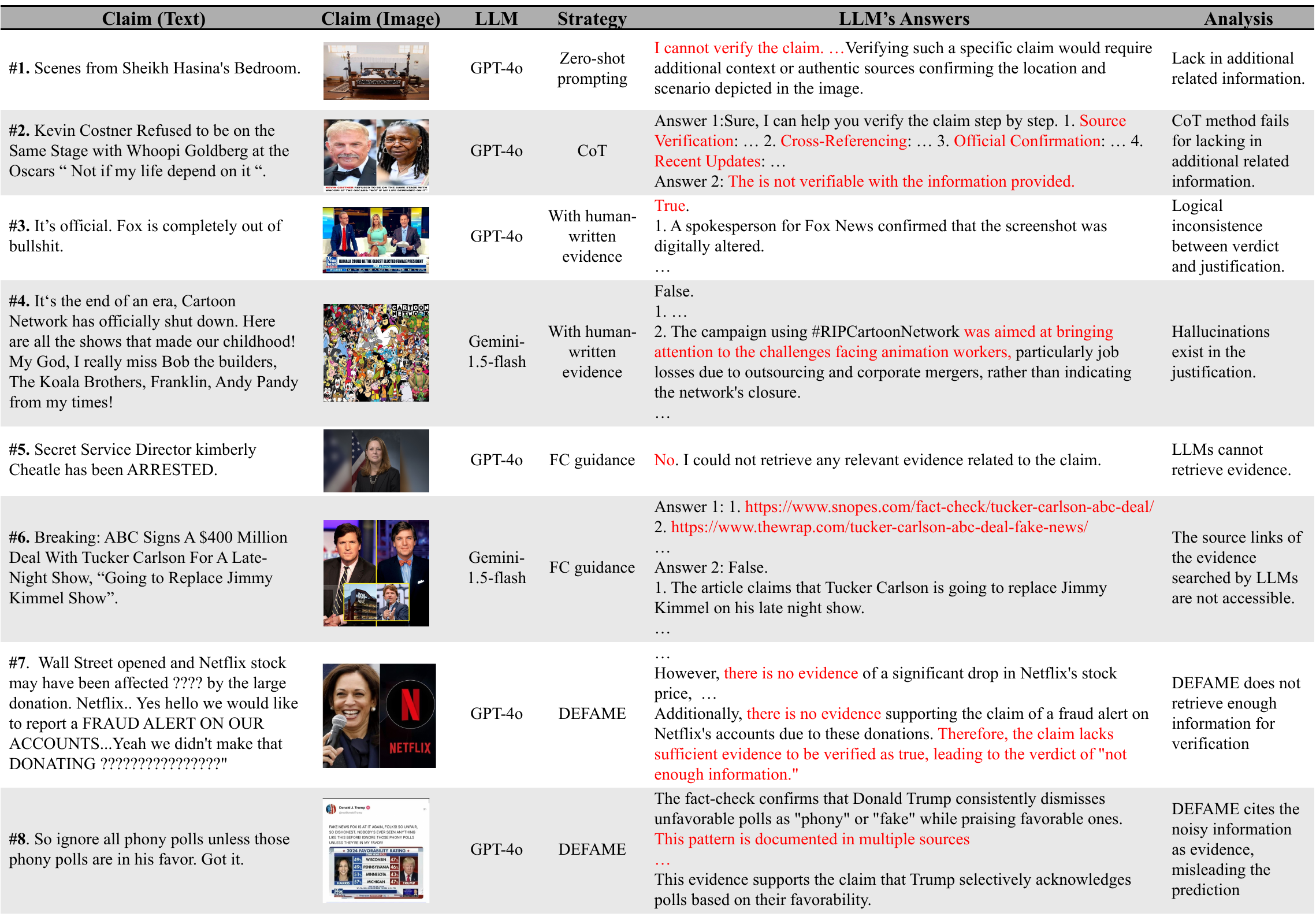}
    \caption{Representative examples in our empirical study. \textit{Analysis} denotes the manual analysis of the failure reason. The incorrect contents of the answers generated by the LLMs are highlighted in red.}
    \label{fig:empirical_study_examples}
\end{figure*}

The experimental results are shown in Table~\ref{tab:empirical-study-1} and some representative examples are shown in Figure~\ref{fig:empirical_study_examples}. The correctness rates of these two LLMs under zero-shot or CoT strategies are very close, both lower than 50\%, indicating that neither of them can complete the task effectively. Compared with zero-shot prompting, the verification rates of the two LLMs slightly increase under CoT. This can be attributed to the fact that CoT guides LLMs to generate structured reasoning steps, enhancing their confidence in attempting verification. However, both approaches are constrained by the same closed-world knowledge boundary and lack access to external evidence, limiting their performance.

We further analyze the LLM responses to understand why they sometimes fail to complete the verification task. The LLMs utilize the CoT approach to generate a series of general steps aimed at verifying the claim from various perspectives. However, they often struggle to access specific information relevant to these steps and the target claim. We present two concrete examples in Figure~\ref{fig:empirical_study_examples}. In \textit{case \#1}, GPT-4o aims to validate the claim using its own text and image through zero-shot prompting. However, it answers that it cannot verify the claim without additional information. In \textit{case \#2}, GPT-4o generates 4 steps to verify the claim through CoT, as shown in answer 1. However, it fails because it does not find additional information related to these 4 aspects, as shown in answer 2. 

In summary, due to the lack of sufficient context and specific information, LLMs cannot generate accurate judgments on the claim independently. Consequently, they frequently fail to verify the truthfulness of disinformation, particularly for claims that have emerged after the cutoff date of their training data. This highlights a critical challenge in relying solely on LLMs for disinformation verification in dynamically evolving information environments.

\begin{tcolorbox}[colback=gray!25!white, colframe=black, boxrule=0.5pt, size=title, breakable,boxsep=1mm, before={\vskip1mm}, after={\vskip0mm}]
\textbf{Finding 1:} \textit{LLMs CANNOT accurately verify the truthfulness of the claim beyond their knowledge cutoff date directly.} 
\end{tcolorbox}

\subsection{(RQ2) Verification with Fact-checking Guidance}

We evaluate whether human-provided fact-checking guidance can improve LLMs' disinformation detection performance. The results are shown in Table~\ref{tab:empirical-study-1} (“FC guidance” row). We then manually analyze the verdicts and justifications generated.

\noindent \textbf{Verdict Analysis.} With fact-checking guidance, the performance gap between GPT-4o and Gemini-1.5-flash widens. GPT-4o shows a high correctness rate (81.8\%) but a low verification rate (20.4\%), often refusing to verify claims due to lack of external access (e.g., Figure~\ref{fig:empirical_study_examples}, case \#5). This means only a small fraction of claims are correctly verified. In contrast, Gemini-1.5-flash achieves a much higher verification rate (91.2\%) but a slightly lower correctness rate (78.6\%). However, many of its seemingly correct predictions rely on unsupported or fabricated evidence, as detailed in the justification analysis. Both models perform better on true claims than on false ones.

\noindent \textbf{Justification analysis.} To assess the reliability of the models’ verdicts, we examine the justifications generated during fact-checking. Ideally, these justifications should include supporting evidence and corresponding source links. Among the correctly verified claims, GPT-4o provided source links in 45.5\% of cases, while Gemini-1.5-flash did so in 50.0\% of cases (e.g., \textit{case \#6} in Figure~\ref{fig:empirical_study_examples}, answer 1). However, nearly all of these links were inaccessible or invalid, and those reachable links were often irrelevant to the claim. Because both LLMs lack real-time web access and the fact-checking samples were published after their training cut-off dates, these references are likely hallucinated. Thus, even when the model outputs a correct verdict, it may arrive at the answer by coincidence rather than by consulting verifiable evidence. In practical fact-checking scenarios, where the ground-truth label is unknown, such unverifiable or fabricated references make the verdicts untrustworthy and raise significant concerns about reliability.

Based on the above analysis, we conclude that LLMs are unable to retrieve evidence from the Internet, and the evidence can be potentially generated by their hallucinations with inaccessible source links. This undermines users' trust in applying LLMs to disinformation detection. 

\begin{tcolorbox}[colback=gray!25!white, colframe=black, boxrule=0.5pt, size=title,breakable,boxsep=1mm,before={\vskip1mm}, after={\vskip0mm}]
\textbf{Finding 2:} \textit{LLMs have shortcomings in \textbf{searching for claim-relevant public information} and their responses may include hallucinated links that weaken result trustworthiness.}
\end{tcolorbox}

\subsection{(RQ3) Verification with Evidence}
\label{subsec:verification_with_evidence}

We evaluate how different sources of external evidence affect LLM-based claim verification: human-written evidence from expert and automatically retrieved evidence produced by an agent-based framework (DEFAME). The results presented in Table~\ref{tab:empirical-study-1}(''Human-written evidence'' and ''DEFAME'' row, respectively). Providing LLMs with external evidence can actually improve their verification performance. However, a clear performance gap emerges across different evidence sources. When supplied with human-authored expert evidence, GPT-4o achieves a correctness rate of 90.3\%. In contrast, GPT-4o attains a lower correctness rate of 63.7\% when using evidence from the automated retrieval framework DEFAME\footnote{We only use GPT-4o as backbone model for DEFAME, as Gemini models are not provided in this work}.

\noindent \textbf{Failure case analysis with automatically retrieved evidence (DEFAME).} Failure cases under this settings reveals two dominant error patterns. 
The first failure type is caused by \textit{evidence insufficiency}. The automated retrieval framework fails to collect sufficiently precise or complete evidence to verify the claim, leading the model cannot decide the claim is true of false. In the generated justifications, the model output as \emph{``the fact-checking process did not find sufficient evidence''}(Figure~\ref{fig:llm_search_failure_case}, \textit{case \#7}) This indicates that the retrieved evidence does not adequately cover the key factual factors required for verification, rather than the claim being inherently unverifiable.
The second failure type arises from \textit{noisy or weakly relevant evidence}, which can actively mislead the verification process and result in incorrect verdicts. In such cases, the justification cites factually correct but irrelevant information. In particular, the evidence differs in stance, scope, or verification target from ground-truth human-written evidence. 
As a result, the model relies on this tangential information as affirmative evidence and produces a True verdict, where the claim is actually false(Figure~\ref{fig:llm_search_failure_case}, \textit{case \#8}).

\noindent \textbf{Failure case analysis with human-written evidence.} We manually examined the errors and identified two main types of failure. The first and more prominent issue is a mismatch between the verdict and the justification. In these cases, the model generates a justification that correctly interprets the evidence and aligns with the ground truth (e.g., citing evidence that refutes a false claim), but the final verdict contradicts it (e.g., predicting ``true''). For example, in \textit{Case \#3} (Figure~\ref{fig:empirical_study_examples}), the claim is labeled false, and GPT-4o provides reasoning that clearly refutes it, yet the model’s final output is ``true.'' This behavior suggests that the reasoning and classification components within the model may be loosely coupled: the model can summarize evidence accurately but fails to map that reasoning consistently to the correct binary label. Secondly, a less frequent failure involves hallucinations in justifications. We observed one hallucination from GPT-4o and two from Gemini-1.5-flash, where the model introduced fabricated or distorted information not present in the evidence. For example, in \textit{Case \#4} (Figure~\ref{fig:empirical_study_examples}), Gemini-1.5-flash added spurious details (highlighted in red), leading to a justification inconsistent with the provided evidence. These errors appear to stem from models’ tendency to overgeneralize or misinterpret evidence.

From the above analysis, we conclude that the effectiveness of LLM-based fact-checking critically depends on the quality of the evidence. When LLMs are supported by high-quality human-written evidence, verification errors are rare. In such cases, LLMs are able to produce coherent and faithful justifications by accurately summarizing the evidence to support their decisions. In contrast, failures primarily arise when evidence is insufficient, noisy, or misaligned. These observations highlight that high-quality evidence not only improves verification accuracy but also enhances the interpretability and trustworthiness of LLM-generated justifications.

\begin{tcolorbox}[colback=gray!25!white, colframe=black, boxrule=0.5pt, size=title,breakable,boxsep=1mm, before={\vskip1mm}, after={\vskip0mm}]
\textbf{Finding 3:} \textit{Evidence quality is a decisive factor in LLM-based multimodal fact-checking.}
\end{tcolorbox}

\section{An Illustrative Example of \Name}
\label{sec:illustrative_example}

\begin{figure*}[t]
    \centering
    \includegraphics[width=1.0\linewidth]{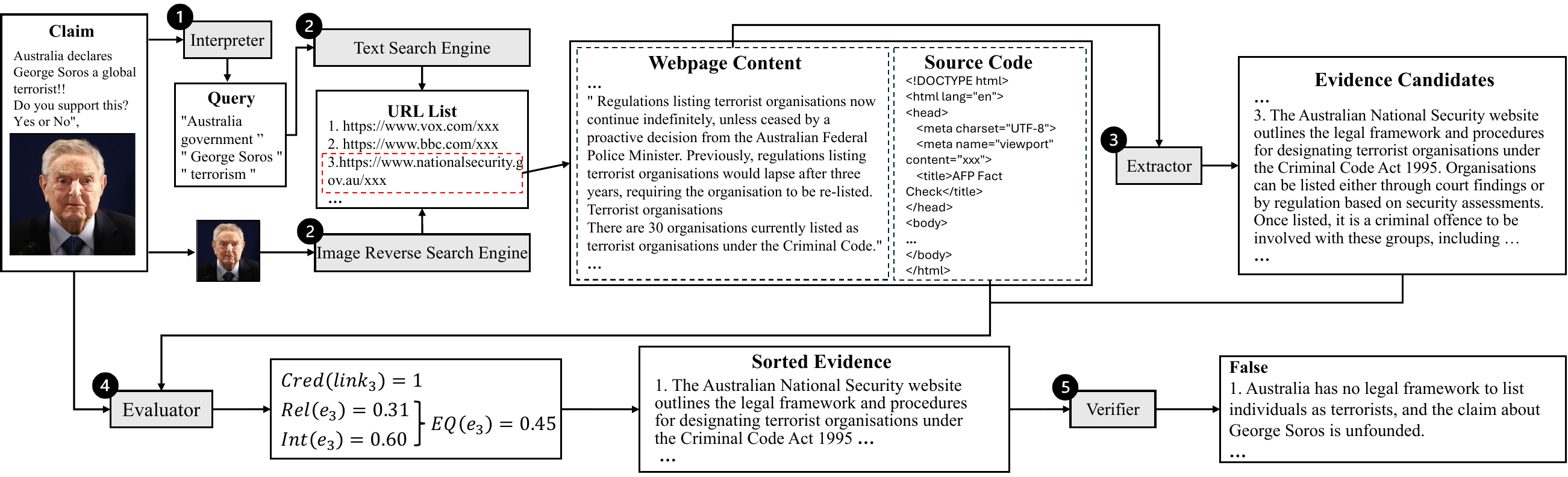}
    \caption{An illustrative example of how \Name automatically verifies the real-world multimodal misinformation.}
    \label{fig:illustrative_example}
\end{figure*}

We use a real-world disinformation example to illuminate how \Name automatically checks the claim's truthfulness, as shown in Figure~\ref{fig:illustrative_example}. The claim states that \textit{"Australia declares George Soros a global terrorist!! Do you support this? Yes or No"} with a portrait of George Soros. The fact-checking process has five stages. \ding{182} \Name first comprehends the multimodal semantics of the claim, encompassing both text and image. Based on this understanding, it generates a query ("Australia government" "George Soros" "terrorism") that is subsequently used for evidence retrieval in the next stage.  \ding{183} \Name searches the sources of the query and image of the claim on the Internet, respectively, and obtains 20 links in total. \ding{184} Then \Name crawls the webpage content and the source code of these links and summarizes the precise abstract of the webpage content. These 20 summaries are evidence candidates that await quality evaluation. We take candidate 3 whose source is link 3 (\textit{https://www.nationalsecurity.gov.au/xxx}) as an example. Its main content is that \textit{"The Australian National Security website outlines the legal framework and procedures for designating terrorist organisations under the Criminal Code Act 1995"}. \ding{185} The credibility value of link 3 is 1, indicating it is credible. The evidence quality score of candidate 3 is 0.45, which is the highest among all the candidates. \ding{186} \Name uses the top 5 pieces of evidence in the sorted evidence set to verify the claim and outputs the verdict and the justification. \Name verifies the claim as false. The justification generated from the example evidence is \textit{"Australia has no legal framework to list individuals as terrorists, and the claim about George Soros is unfounded."}.

\section{Detailed Experiment Settings}
\label{sec:detailed_settings}

\subsection{Settings}
\label{subsec:settings}

\noindent \textbf{Benchmark Datasets.} This appendix provides detailed statistics and construction procedures for the datasets used in our evaluation. For \textbf{RQ1}, we evaluate \Name on two widely used multimodal disinformation benchmarks, Mocheg~\cite{related-work-afc-3} and MR2~\cite{related-work-afc-2}. Mocheg consists of textual claims accompanied by multimodal supporting evidence and labels. MR2 contains multimodal claims, multimodal evidence, and corresponding labels. We remove samples labeled as \emph{Not Enough Information (NEI)} in Mocheg and \emph{unverified} samples in MR2, as our verification task is formulated as binary classification. The dataset statistics, including training and test splits, are summarized in Table~\ref{tab:evaluation_datasets}.

\noindent \textbf{MMDV Dataset Construction.} To support \textbf{RQ2} and \textbf{RQ3}, we construct a new dataset, \textit{MMDV} (Multi-Source Multimodal Disinformation Verification Dataset), designed for evaluating end-to-end, open-world verification.
Unlike existing benchmarks, MMDV provides only multimodal claims and ground-truth labels, without any predefined supporting evidence.
MMDV is constructed to satisfy two key requirements:
(1) all samples contain only claims and labels, requiring models to autonomously retrieve evidence during verification;
(2) all claims are published after the knowledge cutoff dates of the evaluated LLMs (e.g., GPT-4o and Gemini-1.5-Flash), preventing models from relying on memorized knowledge.

We collect samples from three professional fact-checking organizations: Snopes\footnote{\url{https://www.snopes.com/}}, PolitiFact\footnote{\url{https://www.politifact.com/}}, and Reuters.
For Snopes and PolitiFact, we extract textual claims, associated images, and verdicts from fact-check articles, and map their fine-grained labels to binary labels (true/false) following Table~\ref{tab:complete_labels_map}.
For Reuters, we follow the same collection strategy described in Section~\ref{subsubsec:rq1-dataset-construction}.
The final MMDV dataset contains 1,214 multimodal claims, with a balanced distribution of true and false labels. Table~\ref{tab:evaluation_datasets} summarizes the statistics of all datasets used in our experiments, including the number of positive and negative samples and corresponding splits.

\begin{table}[t]
    \caption{Sample statistics of the benchmarks. Positive samples and negative samples denote true and false information, respectively. \ding{55} indicates the benchmark does not have this set.}
    \centering
    \resizebox{\linewidth}{!}{
    \begin{tabular}{cccccc}
    \toprule
         & \multicolumn{2}{c}{\# Positive Samples} & \multicolumn{2}{c}{\# Negative Samples} & \multirow{2}{*}{Lables} \\ \cmidrule{2-5}
         & Train & Test & Train & Test & \\
         \midrule
    Mocheg     & 3,826 & 817 & 4,542 & 825 & supported, refuted \\
    MR2     & 1,854 & 411 & 1,134 & 391 & non-rumor, rumor\\
    MMDV    & \ding{55} & 609 & \ding{55} & 605 & true, false\\
    \bottomrule

    \end{tabular}}

    \label{tab:evaluation_datasets}
\end{table}

\noindent \textbf{Baselines.} Here are brief descriptions of the baseline methods used in our experiments.
(1) \textit{End2End.}~\cite{related-work-afc-3} is a multimodal fact-checking framework that verifies claims using evidence retrieved from a manually constructed closed-domain knowledge base. Its shortcoming is not supporting open-domain web search.
(2) \textit{RB.}~\cite{related-work-afc-2} retrieves evidence from open sources and performs multimodal verification. However, it does not include explicit mechanisms for evaluating evidence quality or generating structured justifications.
(3) \textit{Pre-CoFactv2.}~\cite{pre_cofactv2} focuses on multimodal claim verification using pre-collected evidence. It does not support open-domain evidence retrieval or justification generation.
(4) \textit{SpotFakePlus.}~\cite{spotfakeplus} is a multimodal fake news detection model based on feature fusion across modalities. It performs multimodal classification but does not incorporate evidence retrieval or explainable verification.
(5) \textit{DEFAME.}~\cite{defame} is an LLM-based multimodal verification framework that retrieves information from the public web. While it supports open-domain search and multimodal verification, it does not assess evidence quality. The LLM is deployed with the default settings(GPT-4o).

\noindent \textbf{Settings.} Here are the detailed implementation and environment settings of our experiment.
To ensure fair comparison, we replicate all baseline methods using their official implementations and recommended configurations.
We strictly follow the specified versions of Python and third-party dependencies reported in the original papers.
We implement four variants of \Name: GPT-4o, Gemini-1.5-Flash-001, Llama-3.2-Vision-11B, and Qwen-Vision-7B as both the evidence extractor and the claim verifier.
For each variant, the same LLM backbone is used consistently across all stages of the framework.
Commercial LLMs are accessed via official API endpoints. Open-source models are deployed locally.
All experiments are conducted on Ubuntu 18.04.6 LTS.
Open-source LLMs are deployed on NVIDIA GeForce RTX A6000 GPUs with 48GB VRAM. Baseline models are executed on NVIDIA GeForce RTX 3090 GPUs with 24GB VRAM.
Unless otherwise specified, the temperature parameter of all LLMs is set to zero to reduce output variance and improve reproducibility.

\begin{table}
    \caption{The verdict label mapping used in this paper, which is collected from fact-check agents.}
    \centering
    \resizebox{\linewidth}{!}{
    \begin{tabular}{cc}
    \toprule
    \textbf{Standard Labels} & \textbf{Fact Check Agent Labels} \\ 
    \midrule
    True     &  \makecell[c]{Accurate, Mostly-Accurate, Correct, Partially-Correct, \\Mostly correct, Partially True, Mostly True, True}\\ 
    \midrule
    False     &   \makecell[c]{Misleading, Missing Context, Altered, Synthetic Media, \\Miscapthioned, Satire, Fake News, Inaccurate, Incorrect, \\Likely False, Misrepresented, Missing Context, Mostly False} \\
    \bottomrule
    \end{tabular}}
    \label{tab:complete_labels_map}
\end{table}

\subsection{Parameters Justification}
\label{subsec:parameters_justification}

To explore the optimized hyperparameter \(\alpha\) in Section~\ref{subsubsec:evaluating_evidence}, we set a group of \(\alpha\) with different values $(0.4, 0.5, 0.6)$ and evaluate the performance of \Name in RQ2 in the MMDV dataset with four metrics, respectively. The \Name is deployed with GPT-4o, Gemini-1.5-flash, Llama 3.2-vision, and Qwen-vision. The results are shown in Table~\ref{tab:parameters_justification}, and the best performance of each model among different values of \(\alpha\) is in bold. In general, these four models achieve the best performance, setting \(\alpha\) as $0.5$, indicating a balance between the relevance and integrity of the evidence. The results suggest that relevance and integrity are equally significant when selecting high-quality evidence for fact-checking.

\begin{table}[htbp]
\centering
\caption{Performance of \Name deployed with different LLMs under different value of \(\alpha\).}
\label{tab:parameters_justification}
\resizebox{\linewidth}{!}{
\begin{tabular}{lccccc}
    \toprule
    \textbf{Model} & \textbf{\(\alpha\)} & \textbf{Accuracy} & \textbf{Precision} & \textbf{Recall} & \textbf{F1} \\
    \toprule
    \multirow{3}{*}{GPT-4o} & 0.4 & 88.4\% & 89.1\% & 87.9\% & 88.5\% \\
                               & 0.5 & \textbf{90.2\%} & \textbf{89.9\%} & \textbf{90.0\%} & \textbf{89.8\%}  \\
                               & 0.6 & 87.1\% & 88.2\% & 85.9\% & 87.1\% \\
    \hline
    \multirow{3}{*}{Gemini-1.5-flash} & 0.4 & 84.7\% & 86.3\% & 84.0\% & 85.1\% \\
                               & 0.5 & \textbf{87.0\%} & \textbf{87.8\%} & \textbf{86.7\%} & \textbf{86.8\%} \\
                               & 0.6 & 84.0\% & 86.0\% & 82.7\% & 84.3\% \\
    \hline
    \multirow{3}{*}{Llama 3.2-vsion} & 0.4 & 71.2\% & 73.7\% & 42.7\% & 54.1\% \\
                               & 0.5  & \textbf{73.4\%} & \textbf{70.8\%} & \textbf{56.0\%} & \textbf{48.1\%}\\
                               & 0.6 & 70.9\% & 73.1\% & 41.2\% & 52.5\% \\
    \hline
    \multirow{3}{*}{Qwen-vision} & 0.4  & 68.2\% & 52.4\% & 41.9\% & 46.4\%\\
                               & 0.5  & \textbf{71.1\%} & \textbf{39.6\%} & \textbf{43.4\%} & \textbf{41.6\%}\\
                               & 0.6 & 68.4\% & 47.9\% & 38.6\% & 42.7\%\\
    \bottomrule
\end{tabular}}
\end{table}

\section{Verification Cost and Efficiency of \Name}
\label{sec:cost}

We consider the cost of utilizing \Name for disinformation verification. The cost of \Name is incurred by invoking commercial APIs (LLMs API and Google API). $ {T}_{total} $ and $ {Cost}_{total} $ denote the total execution time and the total invocation cost, respectively. For open-source LLMs, we only compute the elapsed time($ {T}_{total} $). The computing formulas are shown in Equation~\ref{eq:time} and Equation~\ref{eq:fee}. 

\begin{equation}
\label{eq:time}
{T}_{total}={T}_{retrieve}+{T}_{summarize}+{T}_{verify}
\end{equation}


\begin{align}
\label{eq:fee}
{Cost}_{total} &= {Cost}_{retrieve} + {Cost}_{summarize} \notag \\
&\quad + {Cost}_{verify}
\end{align}

The total execution time ($ {T}_{total} $) and the total invocation cost ($ {Cost}_{total} $) of the verification process consist mainly of the following three parts. Note that the framework initializes two threads to execute text direct search and image reverse search in parallel, rather than sequentially, to save as much time as possible.

\begin{enumerate}[leftmargin=*,itemsep=1pt,topsep=0pt,parsep=1pt]

\item Retrieve evidence (${T}_{retrieve}$, ${Cost}_{retrieve}$): Time and cost of Invoking the Google text direct search engine and the image reverse search engine to search for information related to target claim from the Internet.

\item Summarize main content (${T}_{summarize}$, ${Cost}_{summarize}$): Time and cost of Invoking the LLM API to summarize the main content of the original web page.

\item Verify claims (${T}_{verify}$, ${Cost}_{verify}$): Time and cost of invoking the LLM API to verify the claim using the retrieved evidence.

\end{enumerate}

\begin{table}[t]
  \caption{Time cost (s) per disinformation verification. Three stages are included in this process.}
  \centering
  \resizebox{\linewidth}{!}{
  \begin{tabular}{cccccc}
    \toprule
    & GPT-4o & Gemini-1.5-flash & Llamma 3.2-Vision & Qwen-VL & DEFAME\\ 
    \midrule
    Retrieve & 0.1 & 0.1 & 0.1 & 0.1 & -\\
    Summary & 20.3 & 19.7 & 80.2 & 82.4 & -\\
    Verify & 4.2 & 2.0 & 20.1 & 23.8 & -\\ 
    \midrule
    Total & 24.6 & 21.8 & 100.4 & 106.3 & 51.5\\
    \bottomrule
  \end{tabular}}
  \label{tab:cost}
\end{table}

The time cost of \Name with different LLMs is shown in Table \ref{tab:cost}. Overall, Commercial LLMs (23.2s on average) are faster than open-source LLMs (103.4s). Gemini-1.5-flash has the shortest elapsed time of 21.8s. We compute the fees according to the billing rules according to the vendors' portal websites~\footnote{\url{https://openai.com/api/pricing/}}. GPT-4o incurs a small fee of 0.11 USD for each disinformation verification, whereas Gemini-1.5-flash provides free access. As comparison, DEFAME cost 0.24 USD and 51.5s per claim. The most expensive and most time-consuming step during real-time fact check is the summary, as it requires processing massive amounts of text and image data when summarizing the main content from the original web pages. Compared to professional fact-check agents, which require several hours or days to verify a piece of disinformation on average \cite{manual-fact-check-time-consuming-1,manual-fact-check-time-consuming-2}, our method is extremely cost-effective, which greatly reduces elapsed time. 

\section{Evidence Retrieval Method Comparison}
\label{sec:evidence-retrieval-comparison}

\begin{table*}[t]
  \caption{Experimental results with different evidence retrieval approaches. The first row indicates the evidence retrieval approaches. The best metrics are bold for every LLM with different retrieval approaches.}
  \centering
  \resizebox{\linewidth}{!}{
  \begin{tabular}{c||cccc||cccc}
    \hline
    & \multicolumn{4}{c||}{RB} & \multicolumn{4}{c}{\Name} \\
    & Accuracy & Precision & Recall & F1 & Accuracy & Precision & Recall & F1 \\

    \hline
    \Name(Llama 3.2-vision) & 56.3\% & 56.0\% & 54.9\% & \textbf{53.3\%} & \textbf{73.4\%} & \textbf{70.8\%} & \textbf{56.0\%} & 48.1\% \\
    \Name(Qwen-vision) & 47.2\% &36.4\% & \textbf{45.0\%} & 35.2\% & \textbf{71.1\%} & \textbf{39.6\%} & 43.4\% & \textbf{41.6\%} \\
    \Name(Gemini-1.5-flash)   & 65.3\% & 68.9\% & 66.1\% & 64.0\% & \textbf{87.0\%} & \textbf{87.8\%} & \textbf{86.7\%} & \textbf{86.8\%} \\
    \Name(GPT-4o) & 59.4\% & 65.1\% & 62.1\% & 54.8\% & \textbf{90.2\%} & \textbf{89.9\%} & \textbf{90.0\%} & \textbf{89.8\%} \\

    \hline
  \end{tabular}}
  \label{tab:evidence-retrieval}
\end{table*}

To investigate the effectiveness of our evidence-retrieval approach, we conduct experiments to compare the evidence retrieval method in \Name with that used in the previous study~\cite{related-work-afc-2}. We use RB to indicate this evidence retrieval method. The detail of RB is as follows: It initialize a crawler that first uses Google Reverse Image Search to collect textual evidence by crawling descriptions of similar images. Then the crawler identifies image tags, extracts descriptions from \texttt{<figcaption>} and image-related attributes (e.g., \texttt{<alt>}, \texttt{<caption>}), and compiles non-redundant text snippets from each web page for analysis. Additionally, visual evidence is retrieved using the Google Programmable Search Engine with the text of the post as the query, retaining the top 5 images after filtering disinformation sources. We set the MMDV dataset as the benchmark of this experiment and evaluate \Name deployed with four LLMs: GPT-4o, Gemini-1.5-flash, Llama 3.2-vision, and Qwen-vision.

The comparison results are shown in Table \ref{tab:evidence-retrieval}. Keeping other settings identical but with different evidence retrieval methods, \Name performs better in detecting disinformation when using our evidence retrieval method than when using the RB method. The main difference of the two approaches is that we extract the main content of the original web pages, while the RB method collects partial text in HTML tags as evidence, indicating that our method can obtain more abundant and comprehensive information to help LLMs more accurately verify the disinformation.

\section{LLM Prompt Designs in \Name}
\label{sec:prompt_temp}

This section provides the full set of prompt templates used in \Name. These prompts were designed to instruct LLMs in completing different subtasks during the disinformation verification.

The following template is guiding \Name to comprehend the multimodal claim and generate the sub-claims and queries in Section~\ref{subsec:claim_comprehension}.

\begin{tcolorbox}[colback=gray!25!white, size=title,breakable,boxsep=1mm,colframe=white,before={\vskip1mm}, after={\vskip0mm}]
\label{temp:claim_comprehension}

You are a multimodal misinformation interpreter. Your task is to understand a claim that contains both text and image, and generate structured sub-claims and corresponding retrieval queries.

Input:

Text: {claim}

Image: {image}

Output:

1. Sub-claim: ...

   Query: ...
   
...
\end{tcolorbox}

The following template is guiding \Name to summarize the main content of a webpage in Section~\ref{subsubsec:content_extraction}.

\begin{tcolorbox}[colback=gray!25!white, size=title,breakable,boxsep=1mm,colframe=white,before={\vskip1mm}, after={\vskip0mm}]
\label{temp:event_extract}

Suppose you are a professional fact-checker.

Please summarize the provided article by identifying the people (who), the event (what), the location (where), the time (when), the reason (why), the background of the event, the impact of the event, and the follow-up event. Ensure the summary remains concise and clear.
\end{tcolorbox}

The following template is guiding \Name to initialize a verification task in Section~\ref{subsec:claim-verification}.

\begin{tcolorbox}[colback=gray!25!white, size=title,breakable,boxsep=1mm,colframe=white,before={\vskip1mm}, after={\vskip0mm}]
\label{temp:intialization}

Suppose you are a professional fact-checker. I will give you a claim to verify. The following is the claim. \{text\} denotes the text part of the claim. \{image\} denotes the image part of the claim.

Text: \{text\}

Image: \{image\}

Before I provide you with evidence to verify this claim, do nothing but memorize it.

\end{tcolorbox}

The following template is guiding \Name to upload evidence in Section~\ref{subsec:claim-verification}.

\begin{tcolorbox}[colback=gray!25!white, size=title,breakable,boxsep=1mm,colframe=white,before={\vskip1mm}, after={\vskip0mm}]
\label{temp:upload_evidence}

The following list is the evidence related to the claim. You need to remember it and do nothing until the next instruction.

Text evidence: \{text\_evidence\_list\}

\end{tcolorbox}

The following template is guiding \Name to verify the claim in Section~\ref{subsec:claim-verification}.

\begin{tcolorbox}[colback=gray!25!white, size=title,breakable,boxsep=1mm,colframe=white,before={\vskip1mm}, after={\vskip0mm}]
\label{temp:verify}
Verify the claim based on the evidence that I provided to you. The verdict sets of the claim and the verification principle is shown below.

True verdict set: \{true\_verdict\_set\}. False verdict set: \{false\_verdict\_set\}.

(1) If your verification result is in the true verdict set, the claim is true. (2) If your verification result is in the false verdict set, the claim is false.

Next, give the justification for the verdict result. Output your complete answers in the format of the following template.

\{output\_format\}

\end{tcolorbox}

The following template is guiding \Name to output verification results in an explicit format in Section~\ref{subsec:claim-verification}.

\begin{tcolorbox}[colback=gray!25!white, size=title,breakable,boxsep=1mm,colframe=white,before={\vskip1mm}, after={\vskip0mm}]
\label{temp:output_format}

Verdict: True/False.

Evidence:

1. The evidence \{place\_holder\} supports/refutes the{place\_holder} of the claim.

2. The evidence \{place\_holder\} supports/refutes the{place\_holder} of the claim.

3. \ldots\ldots

Summary: Use a concise sentence to summarize including your prediction and reason.
\end{tcolorbox}

\section{Fact Check with LLMs with search capabilities}
\label{sec:llm-search}

In this section, we detail the experiment setup and results in Section~\ref{subsec:Experimental_Setup}, \textbf{baselines}. Analysis shows the limitations of the LLMs with online search functionality on multimodal fact-checking.

\noindent\textbf{Models.} We selected three state-of-the-art LLMs: GPT-4o-search-preview~\cite{gpt_web_search}, GPT-4o-mini-search-preview~\cite{gpt_web_search}, and Gemini-1.5-flash-search-grounding~\cite{gemini_search_grounding} to evaluate on the fact-checking tasks. Before that, we first introduce two common characteristics of these models that are not perfectly aligned with the requirements of the multimodal disinformation detection task, potentially constraining their performance: (1) All of them only support the single text modality. So they cannot handle the visual modality and cannot reverse search for the image. (2) They are end-to-end, black-box models that cannot customize the search query and lack domain-specific customization (whitelist/blacklist), complicating prevention of such access. GPT series models do not reveal search engine and the queries used for retrieval in their responses, whereas Gemini series models utilize the Google search engine, providing the search queries employed during the retrieval process.

\noindent\textbf{Setup.} We invoke APIs to utilize these models and evaluate them on the MMDV dataset with standard classification metrics: accuracy, precision, recall, and F1-score. The sample in the MMDV dataset that needs to be verified contains text and images. Because the selected LLMs do not support image modality~\cite{gpt_web_search, gemini_search_grounding}. Hence, we utilized only the textual portion of multimodal claims. Models were required to provide a binary verdict (true/false) with coherent justifications. In addition to the baseline configuration, we also introduce an improved setup to address the observed limitations from the baseline experiment results that the LLMs' built-in search engine from retrieving the original source of the claim. We detail this in the paragraph Analysis and propose three approaches to prevent such scenario: (1) \textbf{Zero-shot guidance.} This directly instructs LLMs not to retrieve evidence from specific domains. (2) \textbf{Multi-turn conversation.} This guides the LLMs to exclude specific domains that are retrieved in the first turn and regenerate the answers. (3) \textbf{Insert dorks.} This append dorks after the claim, aiming to exclude specific domains when the LLMs are searching online information. The Gemini with the searching tool does not provide a multi-turn conversation; thus, we do not evaluate it under this setup.

\noindent\textbf{Results.} As shown in Table~\ref{tab:search_LLM_performance}, GPT-4o-search-preview achieves the highest accuracy of 88.3\% among these 3 models. Gemini-1.5-flash-search-grounding achieves the lowest accuracy of 79.7\%. Compared to \Name(GPT-4o), GPT-4o-search-preview performs slightly worse, with 1.8\% lower accuracy. The gap is larger for Gemini-1.5-flash: \Name(Gemini-1.5-flash) outperforms its search-grounding version by 7.3\%. The performance of the search-enabled LLMs under the zero-shot guidance, multi-turn conversation, and inserting dorks setting is close to their baseline configuration.

\begin{figure}[t]
    \centering
    \includegraphics[width=\linewidth]{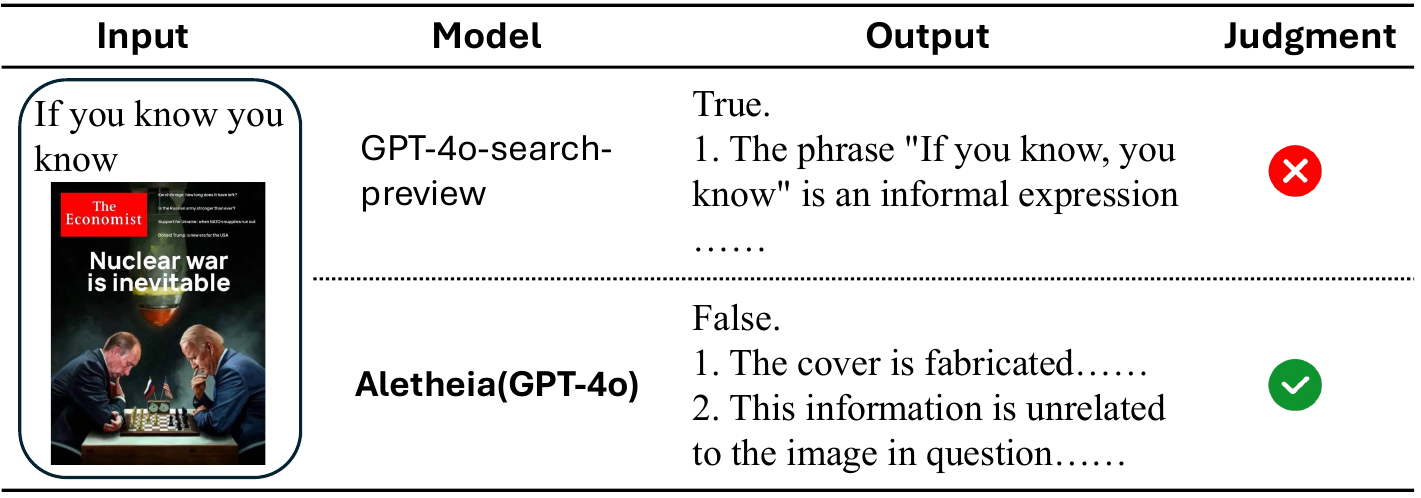}
    \caption{An illustrative example of how GPT-4o-search-preview fails to verify the multimodal claim.}
    \label{fig:llm_search_failure_case}
\end{figure}

\noindent \textbf{Analysis.} We manually check the verification results to explore what leads the models to achieve such high performance and conclude 2 insightful findings: (1) Since the MMDV dataset derives from publicly available fact-checking outcomes, LLMs' embedded search engines inadvertently retrieve these existing results, leading the LLMs to access the ground truth labels before final prediction and achieving high accuracy. Although we employ mitigation to prevent such occurrences,  the results show that these approaches are not effective. This is unfair to compare the performance between these models and \Name. (2) Only using the textual part of a multimodal claim to verify fails to leverage the complementary information present in image modalities. This may overlook modality-specific cues that are critical for accurate fact verification, leading to incomplete or biased verification outcomes. Specifically, our analysis of failure cases reveals two common types of claims where performance drops significantly: 1) claims presented only by images; 2) claims with textual and visual content, while the key information is conveyed mainly through visual content. Because these models are unable to retrieve or process visual evidence, they often fail to verify such claims.

\noindent \textbf{Case.} The illustrative example in Figure~\ref{fig:llm_search_failure_case} demonstrates how GPT-4o-search-preview fails but \Name succeeds to verify the multimodal claim. The claim consists of text(\textit{if you know you know} and an image(a magazine cover that includes Putin and Trump), whose ground-truth label is false. GPT-4o-search-preview failed to verify the claim because it only relied on text for retrieval. However, the full meaning of the claim depended on both the text and the image. As a result, the retrieved evidence from only text is unrelated to the full semantics of the multimodal claim. This indicates the limitation of these models for multimodal disinformation fact-checking.

\begin{table}[t]
  \caption{The experimental results for LLMs with search function on MMDV dataset. Zero-shot, Multi-turn, and dorks indicate the experiment settings: zero-shot guidance, multi-turn conversation, and inserting dorks, respectively(in the paragraph: Setup).}
  \centering
  \resizebox{\linewidth}{!}{
  \begin{tabular}{c||cccc}
    \hline
    & Accuracy & Precision & Recall & F1 \\

    \hline
    Gemini-1.5-flash-search-grounding & 79.7\% & 80.9\% & 79.8\% & 79.5\% \\
    GPT-4o-search-preview & 88.3\% & 88.4\% & 88.3\% & 88.2\%\\
    GPT-4o-mini-search-preview & 85.3\% & 86.5\% & 85.5\% & 85.3\%\\
    \midrule
    Gemini-1.5-flash-search-grounding(zero-shot) & 77.3\% & 78.6\% & 77.4\% & 77.1\% \\
    GPT-4o-search-preview(zero-shot) & 88.1\% & 88.4\% & 87.9\% & 87.9\%\\
    GPT-4o-mini-search-preview(zero-shot) & 84.9\% & 84.8\% & 85.0\% & 84.9\%\\
    \midrule
    Gemini-1.5-flash-search-grounding(Multi-turn) & - & - & - & - \\
    GPT-4o-search-preview(Multi-turn) & 87.8\% & 87.9\% & 87.7\% & 87.8\%\\
    GPT-4o-mini-search-preview(Multi-turn) & 84.8\% & 85.1\% & 84.6\% & 84.9\%\\
    \midrule
    Gemini-1.5-flash-search-grounding(dorks) & 77.8\% & 79.6\% & 77.9\% & 77.5\% \\
    GPT-4o-search-preview(dorks) & 87.5\% & 87.3\% & 87.9\% & 87.6\%\\
    GPT-4o-mini-search-preview(dorks) & 85.1\% & 85.5\% & 84.9\% & 85.1\%\\
    \midrule
    \Name (Gemini-1.5-flash) & 87.0\% & 87.8\% & 86.7\% & 86.8\% \\
    \Name (GPT-4o) & \textbf{90.2\%} & \textbf{89.9\%} & \textbf{90.0\%} & \textbf{89.8\%} \\
    \hline
  \end{tabular}}
  \label{tab:search_LLM_performance}
\end{table}

\noindent\textbf{Conclusion.} Due to the above drawbacks, despite SOTA LLMs equipped with search tools achieving relatively high accuracy, the results are misleading. This is because they verify claims under unrealistic scenarios where they can access ground-truth content during the verification process, rather than performing multimodal fact-checking as required in real-world settings. Hence, they are not mature enough to be applied to the multimodal fact check and need further improvement. In contrast, \Name is specifically designed to address the challenges of multimodal disinformation. It incorporates image reverse search tools to retrieve evidence relevant to the visual content, enabling it to capture key information that text-only systems overlook. Consequently, \Name demonstrates greater robustness and reliability in verifying complex multimodal claims.



\end{document}